\documentclass{article}

\PassOptionsToPackage{numbers, compress}{natbib}

\usepackage[preprint]{neurips_2024}

\usepackage[utf8]{inputenc} 
\usepackage[T1]{fontenc}    
\usepackage{hyperref}       
\usepackage{url}            
\usepackage{comment}
\usepackage{booktabs}       
\usepackage{amsfonts}       
\usepackage{nicefrac}       
\usepackage{microtype}      
\usepackage{amsmath}
\usepackage{amsthm}
\usepackage{booktabs}
\usepackage{algorithm}
\usepackage{algorithmic}
\usepackage{amssymb}
\usepackage{graphicx,wrapfig}
\usepackage{booktabs}
\usepackage{multirow}
\usepackage{bbm}
\usepackage[table,xcdraw,dvipsnames]{xcolor}
\usepackage{marvosym}
\usepackage{listings}  
\usepackage{mylstlinebgrd}
\newtheorem{conclusion}{Core Conclusion}
\usepackage{cleveref}
\crefrangeformat{table}{Tables~#3#1#4--#5#2#6}

\definecolor{eclipseBlue}{RGB}{42,0.0,255}
\definecolor{eclipseGreen}{RGB}{63,180,95}
\definecolor{eclipsePurple}{RGB}{175,0,25}
\definecolor{codewhite}{rgb}{0.70,0.70,0.70}
\definecolor{codegray}{rgb}{0.5,0.5,0.5}
\definecolor{codepurple}{rgb}{0.58,0,0.82}
\definecolor{backcolour}{rgb}{0.95,0.95,0.92}

\definecolor{instructioncolor}{rgb}{0.6275, 0.7686, 0.6157}
\definecolor{taskcolor}{rgb}{0.8824, 0.9255, 0.7843}
\definecolor{prefixcolor}{rgb}{0.9922, 1, 0.6824}

\lstdefinelanguage{Prompt}{
	backgroundcolor=\color{backcolour},   
	keywordstyle=\color{magenta},
	numberstyle=\tiny\color{codegray},
	basicstyle=\ttfamily\footnotesize,
	breakatwhitespace=false,         
	breaklines=true,   
    breakindent=-5pt,
	captionpos=b,                    
	keepspaces=true,                 
	numbers=left,                    
	numbersep=5pt,                  
	showspaces=false,                
	showstringspaces=false,
	showtabs=false,                  
	tabsize=4,
	escapeinside={`}{`},
	morecomment = [s][\color{eclipseGreen}\bfseries]{How}{?},
        morecomment = [l][\color{eclipseBlue}\bfseries]{SELECT},
        morecomment = [l][\color{eclipsePurple}\bfseries]{\$\{DATABASE_SCHEMA\}},
        morecomment = [s][\color{eclipsePurple}\bfseries]{CREATE}{;},
        morecomment = [l][\color{eclipsePurple}\bfseries]{Table},
        morecomment = [l][\color{eclipsePurple}\bfseries]{stadium},
        morecomment = [l][\color{eclipsePurple}\bfseries]{singer},
        morecomment = [l][\color{eclipsePurple}\bfseries]{concert},
        morecomment = [l][\color{eclipsePurple}\bfseries]{singer_in_concert},
        morecomment = [l][\color{codewhite}\bfseries]{\$\{TARGET_QUESTION\}},
    postbreak={
       \mbox{
           \lst@linebreakbgrd
           \rotatebox[y=0.7ex]{180}{\color{black}$\Lsh\,$}
       }
    },
}

\lstdefinelanguage{Special_Prompt}{
	backgroundcolor=\color{backcolour},   
	keywordstyle=\color{magenta},
	numberstyle=\tiny\color{codegray},
	basicstyle=\ttfamily\footnotesize,
	breakatwhitespace=false,         
	breaklines=true,   
    breakindent=-5pt,
	captionpos=b,                    
	keepspaces=true,                 
	numbers=left,                    
	numbersep=5pt,                  
	showspaces=false,                
	showstringspaces=false,
	showtabs=false,                  
	tabsize=4,
	escapeinside={`}{`},
        moredelim=**[is][\color{eclipsePurple}\bfseries]{\&\&}{\&\&},
        moredelim=**[is][\color{eclipseGreen}\bfseries]{\^\^}{\^\^},
        moredelim=**[is][\color{eclipseBlue}\bfseries]{@@}{@@}, 
    postbreak={
       \mbox{
           \lst@linebreakbgrd
           \rotatebox[y=0.7ex]{180}{\color{black}$\Lsh\,$}
       }
    },
}

\lstdefinelanguage{Question}{
	backgroundcolor=\color{backcolour},   
	sensitive = true,
	morecomment = [s]{Gold}{SQL},
	commentstyle ={\color{red}\bfseries\underbar},
	morestring = [b]",
	morestring = [b]',
	stringstyle = \color{eclipseGreen},
	basicstyle=\ttfamily\footnotesize,
	breaklines=true,
	alsoletter=!?-,
	emph={StableVicuna-13B, Vicuna-13B, Gold-SQL, Write a sql to answer},
	emphstyle={\color{red}\bfseries\underbar},
	captionpos=b,
	escapeinside={[}{]},
	keepspaces=true,              
	showspaces=false,                
	showstringspaces=false,
	showtabs=false,                  
	tabsize=4,
	columns=flexible
}

\title{SQLBench: A Comprehensive Evaluation for Text-to-SQL Capabilities of Large Language Models}

\author{%
  Bin Zhang$^{1,2,*}$\quad 
  Yuxiao Ye$^{5,}$\thanks{These authors contributed equally.}\quad 
  Guoqing Du$^{3}$\quad 
  Xiaoru Hu$^{3}$\quad 
  Zhishuai Li$^3$\quad 
  Chi Harold Liu$^5$\and
  \textbf{Zhiwei Xu$^4$\quad
  Guoliang Fan$^{1,2}$\quad
  Rui Zhao$^3$\quad
  Ziyue Li$^{6}$\quad
  Hangyu Mao$^{7,}$\textsuperscript{\Letter}}\\
  \\
  $^1$Institute of Automation, Chinese Academy of Sciences\\
  $^2$School of Artificial Intelligence, University of Chinese Academy of Sciences\\
  $^3$SenseTime Research\\
  $^4$School of Artificial Intelligence, Shandong University\\
  $^5$School of Computer Science and Technology, Beijing Institute of Technology\\
  $^6$Technical University of Munich, Heilbronn Data Science Center, Munich Data Science Institute\\
  $^7$Institute of Microelectronics, Chinese Academy of Sciences
}

\begin{document}

\maketitle

\begin{abstract}
  Large Language Models (LLMs) have emerged as a powerful tool in advancing the Text-to-SQL task, 
significantly outperforming traditional methods.
Nevertheless, as a nascent research field, there is still no consensus on the optimal prompt templates and design frameworks. 
Additionally, existing benchmarks inadequately explore the performance of LLMs across the various sub-tasks of the Text-to-SQL process, 
which hinders the assessment of LLMs' cognitive capabilities and the optimization of LLM-based solutions.
To address the aforementioned issues, we firstly construct a new dataset designed to mitigate the risk of overfitting in LLMs. 
Then we formulate five evaluation tasks to comprehensively assess the performance of diverse methods across various LLMs throughout the Text-to-SQL process.
Our study highlights the performance disparities among LLMs and proposes optimal in-context learning solutions tailored to each task. 
These findings offer valuable insights for facilitating the development of LLM-based Text-to-SQL systems.
\end{abstract}

\section{Introduction}

Text-to-SQL automatically transforms natural language (NL) questions into structured SQL statements, e.g., based on the user's question ``\textit{How many musicians do we have?}'' and the database schema (e.g., table/column names), the ideal Text-to-SQL module can generate SQL statements ``\texttt{SELECT count(*) FROM musician}'', which will then be executed on the SQL database engine to retrieve the correct answer. Text-to-SQL 
facilitates seamless user interaction with databases~\cite{qin2022survey}.
Previous approaches 
primarily focus on pattern matching between natural language and SQL statements, utilizing machine learning models to acquire the mapping between the two sequences~\cite{seq2sql, resdsql}. 
However, 
Large Language Models (LLMs) have brought about a substantial transformation in this field~\cite{c3, din-sql} 
as powerful tools~\cite{ruan2023tptu, kong2023tptu, sui2023reboost}, showcasing tremendous potential in comprehending complex NL questions and generating accurate SQL statements. 
By combining advanced reasoning techniques and in-context learning capabilities, LLMs have significantly pushed the boundaries of the state-of-the-art in this domain, outperforming traditional methods by a considerable margin. 


Despite the continuous improvement of LLM-based methods in the two popular datasets, i.e., Spider~\cite{spider} and BIRD~\cite{bird}, there remains a critical gap in the systematic benchmarking of these solutions~\cite{text2sql_survey2023, qin2022survey, kumar2022survey2}. 
\textbf{(1) Are the current solutions overfitting Spider and BIRD?} The risk of overfitting, particularly for LLMs trained on coding tasks and open-source datasets, poses a significant challenge to the generalizability 
~\cite{codellama}. 
\textbf{(2) What is the optimal prompt?} The optimal prompt engineering strategies 
play a crucial role in guiding LLMs to generate accurate SQL queries. 
Although various design schemes are explored
~\cite{dail-sql}, there is no consensus on the most effective prompt template.
\textbf{(3) Is Text-to-SQL the only task that matters?} Furthermore, the current benchmarks \cite{spider, bird}, while comprehensive in their assessment of end-to-end Text-to-SQL tasks, have not yet provided a detailed exploration of the models' performance across other various sub-tasks and components of the Text-to-SQL process~\cite{pourreza2023evaluating, liu2023comprehensive}.
A detailed exploration of these sub-tasks is crucial for a thorough evaluation of LLMs' cognitive capabilities and their role in facilitating the Text-to-SQL process.
Thus, it is essential to develop a more granular benchmarking approach that can accurately reflect the multifaceted nature of Text-to-SQL and inform 
more effective LLM-based solutions. 

\begin{figure*}
\centering
\includegraphics[width=\textwidth]{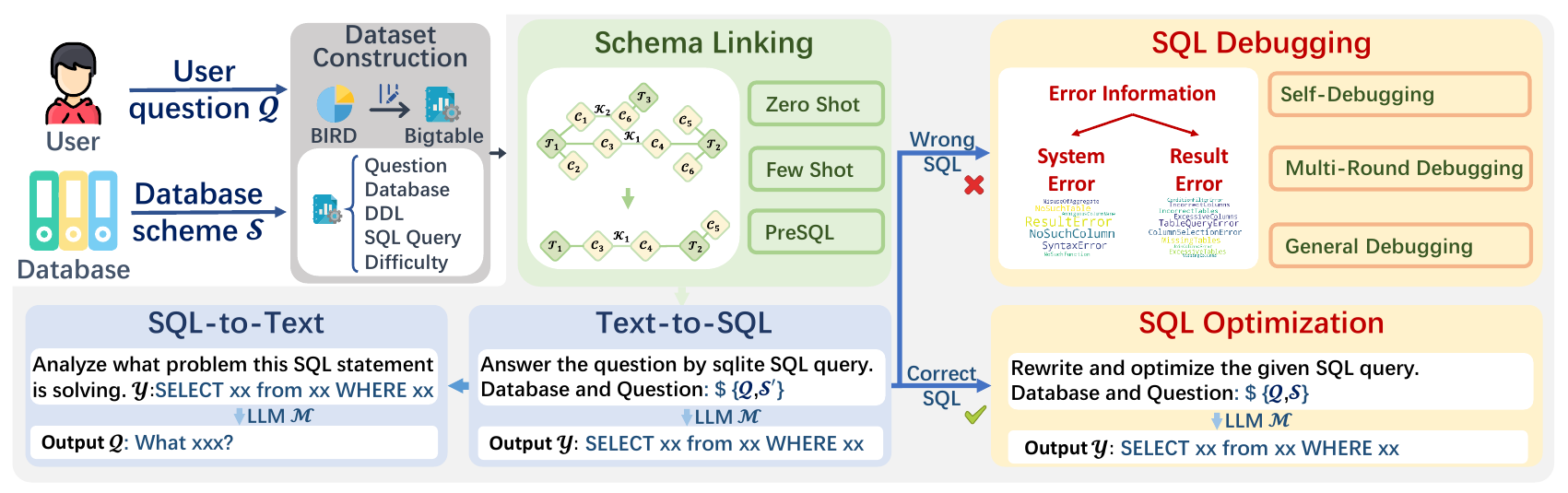}
\vspace{-6mm}
\caption{Benchmarking tasks within SQLBench, encompassing various Text-to-SQL pipeline stages. 
We first construct the Bigtable dataset. Subsequently, different types of LLMs, including general-purpose and coding-specific models, are employed to evaluate various subtasks.
These subtasks include end-to-end Text-to-SQL, SQL-to-Text tasks and Schema Linking, which aligns entity references in the question with the corresponding schema tables or columns. For the questions that the LLMs fail to process correctly, 
a SQL Debugging task is designed with the aim of having the LLMs identify and correct the errors to generate the correct results. A SQL Optimization task is designed for the parts that the LLMs are able to handle correctly, with the goal of improving the execution efficiency of the SQL statements.}
\label{fig:overall_stru}
\end{figure*}

\begin{figure*}[tb]
\centering
\includegraphics[width=\textwidth]{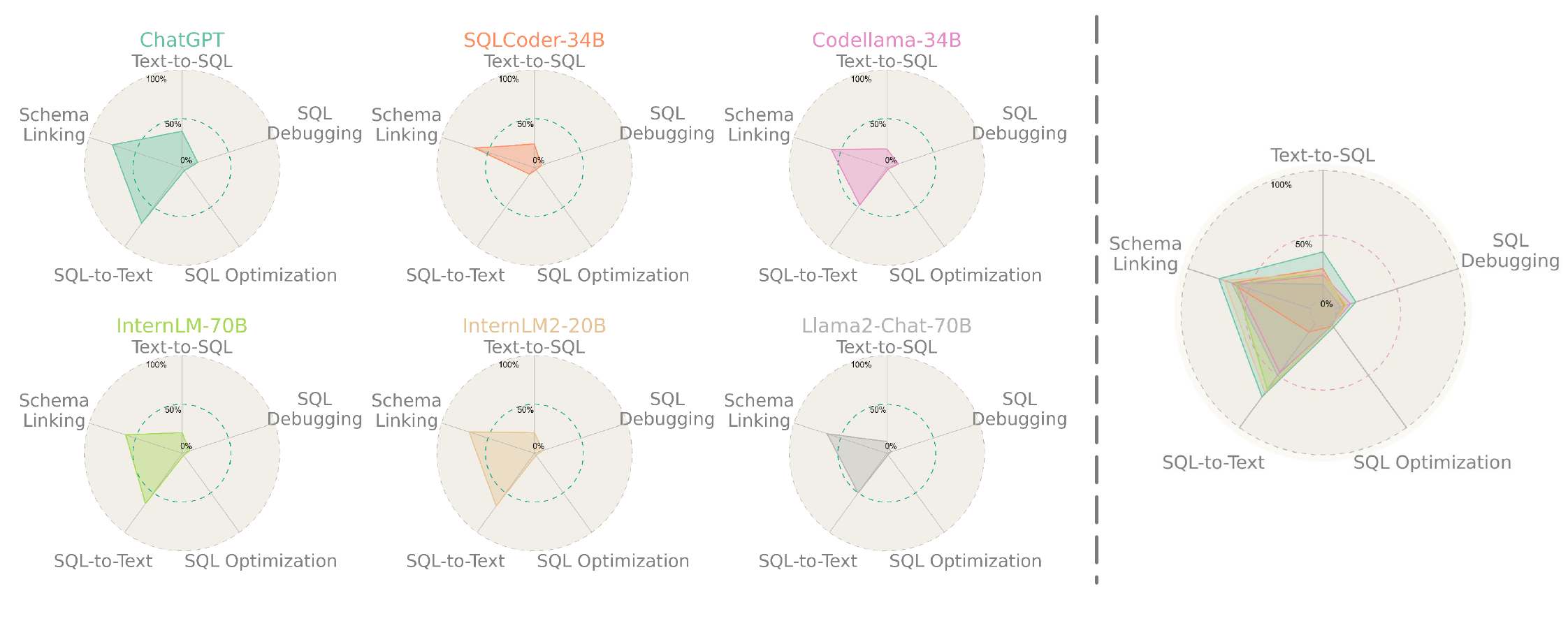}
\vspace{-6mm}
\caption{\textbf{\textit{Left:}} Six LLMs' performance in five sub-tasks of SQLBench.  \textbf{\textit{Right:}} The summarized performance of all LLMs.}
\label{fig:radar_plot}
\end{figure*}

To address the aforementioned challenges and fill the gap in the systematic benchmarking of LLMs in Text-to-SQL, we construct a comprehensive testing benchmark called \textbf{SQLBench} that provides a holistic assessment of LLM capabilities in this domain. 
Our approach begins with the construction of a Text-to-SQL dataset, designed to mitigate the risk of overfitting by considering question complexities, database sizes, and prerequisite knowledge. It contains questions involving queries across varying numbers of tables. 
This dataset, 
constructed on the foundation of existing open-source data,
complemented by existing open-source data, 
serves as a robust platform for evaluating LLMs across a spectrum of tasks.
Furthermore, we devise five distinct tasks: Text-to-SQL, SQL Debugging, SQL Optimization, Schema Linking, and SQL-to-Text, to comprehensively evaluate LLMs' capabilities across the full spectrum of the Text-to-SQL process (see Figure~\ref{fig:overall_stru}). 
Subsequently, we perform an extensive analysis of various techniques that are essential for improving the in-context learning abilities of LLMs and their precision in generating SQL queries. Specifically, our contributions are summarized as follows:
\begin{itemize}
    \item To determine the optimal prompt template, we partition the prompt text into distinct components and perform thorough testing of LLMs' performance on end-to-end Text-to-SQL tasks across all possible combination patterns.
    \item Our benchmarking approach encompasses a range of LLMs, including both general-purpose and coding-specific models with varying parameter sizes. We determine the performance boundaries of these models and identify their performance disparities (see Figure~\ref{fig:radar_plot}). 
    \item For each task, we systematically assess the impact of information granularity on model performance and identify the optimal context learning strategies, such as zero-shot and few-shot, to maximize the performance of the models.
\end{itemize}

\section{Related Work}
\textbf{Text-to-SQL Methods.}
Numerous traditional learning-based Text-to-SQL methods existed before the emergence of LLMs~\cite{seq2sql, sqlnet, hydranet}. 
These methods fall into two categories based on their network architecture: non-seq2seq and seq2seq.
Non-seq2seq methods~\cite{rat-sql, SADGA, s2sql} first utilize a relation-aware self-attention mechanism for encoding, and then use various decoders for generating SQLs, either as abstract syntax trees~\cite{2017syntax-tree, lge-sql} or via slot-filling~\cite{2018coarse2fine, SQLova, ryansql}.
%
%
Seq2seq methods~\cite{BRIDGE, RASAT} directly translate NL questions into SQL queries through transformer-based~\cite{transformer} architectures in an end-to-end manner. These methods obtain competitive performance by fine-tuning rather than training from scratch, with less effort~\cite{T5, graphix-t5, resdsql}. 
However, the highest accuracy achieved by these traditional methods on Spider is 79.9\%, which is still far from being a reliable Text-to-SQL parser~\cite{text2sql_survey2023}.

Given the powerful zero-shot reasoning and domain generalization abilities of LLMs, LLM-based Text-to-SQL methods successively refresh the record on Spider~\cite{li2024pet}.
C3~\cite{c3}, leveraging ChatGPT, achieved 82.3\% accuracy by refining model inputs, biases, and outputs. DIN-SQL~\cite{din-sql} decomposes the task into manageable sub-tasks, reaching 85.3\% accuracy, while DAIL-SQL~\cite{dail-sql} achieves 86.6\% through fine-tuning and in-context learning optimizations in few-shot settings. 
Similarly, other studies investigate the demo selection by synthesising in-domain examples~\cite{chang2023selective} and retrieving question skeletons~\cite{guo2023prompting}.
%
%
Nevertheless, these studies largely rely on OpenAI LLMs without testing other open-source models, nor do they systematically examine sub-task performance of Text-to-SQL.

\textbf{Datasets and Evaluation Metrics.}
The rapid advancement of Text-to-SQL datasets has expanded their scope from single-domain~\cite{atis, geoquery} to cross-domain~\cite{seq2sql, kaggleDBQA}.
Most of the recent Text-to-SQL works are done on \textbf{Spider}~\cite{spider}, recognized as the most challenging cross-domain benchmark featuring multi-table queries, complex SQL clauses, and nested SQLs.
Many variants of Spider are designed to test the adaptability and robustness of Text-to-SQL methods~\cite{spider-real, spider-syn, spider-dk}.
The newest seminal benchmark is \textbf{BIRD}~\cite{bird}.
It is the first to be curated for evaluating SQL execution efficiency in large-scale databases.
In this paper, we construct a novel dataset built upon BIRD and use it in the evaluation (see Section~\ref{sec:dataset-construction}).

We evaluate the accuracy of SQL queries using \textbf{Execution Accuracy (EX)}, which measures whether the result of executing the predicted query matches the gold value. Additionally, \textbf{Valid Efficiency Score (VES)}, an integrated metric that assesses both the accuracy and efficiency of SQL queries, is employed to evaluate SQL optimization in Section~\ref{sec:sql-optimization}.
Refer to Appendix~\ref{app:metrics} for detailed definitions and discussions of evaluation metrics.

\section{Settings}
\textbf{Task Formulation.}\label{sec:task-formulation}
In an LLM-based Text-to-SQL system, LLMs are employed to facilitate the transformation of natural language questions into executable SQL queries. Specifically, Let $\mathcal{Q}$ be a natural language question and $\mathcal{S}$ be the database schema. $\mathcal{S}$ is defined by a tuple $\mathcal{S} = (\mathcal{T}, \mathcal{C}, \mathcal{K})$, where $\mathcal{T}=\{t_1,...,t_m\}$ represents multiple tables, $\mathcal{C}=\{c_1,...,c_n\}$ represents columns, and $\mathcal{K}$ represents foreign key relationships. The goal is to produce a SQL query $\mathcal{Y}$ such that $\mathcal{Y}$ is executable and accurately represents the intent of $\mathcal{Q}$.
Given the prompt template $\mathcal{P}(\mathcal{Q}, \mathcal{S})$, the generation process of the SQL query $\mathcal{Y}$ by an LLM $\mathcal{M}$ can be formally defined as a conditional probability distribution:
\begin{align}\small
    \mathbb{P}_{\mathcal{M}}(\mathcal{Y}|\mathcal{P}(\mathcal{Q}, \mathcal{S})) = \prod_{i=1}^{|\mathcal{Y}|} \mathbb{P}_{\mathcal{M}}(\mathcal{Y}_i | \mathcal{P}(\mathcal{Q}, \mathcal{S}), \mathcal{Y}_{1:i-1}).
\end{align}
Here, LLM autoregressively generates each token, $\mathcal{Y}_i$ denotes the $i$-th token of the SQL query $\mathcal{Y}$, and $|\mathcal{Y}|$ denotes the length of the query $\mathcal{Y}$.

\textbf{Evaluation Models.}
Our benchmarking study evaluates the performance of two distinct categories of LLMs with varying parameter sizes: general-purpose and coding-specific. General-purpose LLMs are designed for versatile text generation and comprehension across diverse domains, trained on extensive internet text datasets. Specifically, ChatGPT (gpt-35-turbo-16k)
, LLaMa2-Chat-70B~\cite{llama2}, InternLM-70B \cite{team2023internlm} 
and InternLM2-20B are selected as the main baseline models. Coding-specific LLMs are fine-tuned and optimized for programming scenarios, excelling in code generation and technical language understanding. In this paper, the performance analysis of Codellama-34B~\cite{codellama} and SQLCoder-34B \cite{SQLCoder} 
are provided.

\section{Dataset Construction}\label{sec:dataset-construction}

\begin{table}[htbp]
\centering
\scriptsize

\caption{EX (\%) of different LLMs on open source datasets.}
\begin{tabular}{@{}lcccc@{}}
\toprule
 &
  {  \textbf{SQLCoder-34B}} &
  {  \textbf{InternLM-70B}} &
  {  \textbf{Codellama-34B}} &
  \multicolumn{1}{l}{\textbf{LLama2-Chat-70B}} \\ \midrule
{  \textbf{Spider Dev}} &
  65.00 &
  67.40 &
  \textbf{71.60} &
  54.70 \\
{  \textbf{BIRD Dev}} &
  \textbf{32.07} &
  29.60 &
  28.29 &
  20.60 \\ \bottomrule
\end{tabular}%
\label{tab:overfit}
\vspace{-5pt}
\end{table}

We conduct a preliminary assessment of the performance of various LLMs on multiple open-source datasets. As depicted in Table~\ref{tab:overfit}, the performance of LLMs varies inconsistently across different datasets. Specifically, on the Spider dataset, Codellama-34B outperforms InternLM-70B and SQLCoder-34B, while on the Bird dataset, SQLCoder-34B surpasses InternLM-70B and Codellama-34B.
The performance differences between LLMs may stem from variations in the problem types they are adept at handling. Furthermore, as LLMs are trained on extensive datasets, the discrepancies could be attributed to the use of open-source datasets during fine-tuning of coding-specialized LLMs.
This poses challenges in ensuring the reliability of evaluation results obtained on these datasets.

To address the potential overfitting of LLMs, particularly those specialized in coding tasks, and to ensure a reliable and accurate assessment of their capabilities, we construct a novel dataset, termed ``BigTable''. This dataset is an extension and augmentation of the BIRD dataset.
\begin{table}[htbp]
\centering
\scriptsize
\caption{The data distribution of ``BigTable'' includes the average number of ground truth (GT) tables and columns involved in the instances.}
\label{tab:bigtable}
\resizebox{0.65\textwidth}{!}{%
\begin{tabular}{@{}ccc@{}}
\toprule
 { \textbf{No. of GT Tables}} &
  { \textbf{No. of Instances}} &
  { \textbf{Avg. No. of Columns}} \\ \midrule
{ \textbf{1}} &
  50 &
  23.30 \\
  { \textbf{2}} &
  50 &
  56.94  \\
  { \textbf{3}} &
  50 &
  31.73  \\
{ \textbf{>3}} &
  50 &
  29.04  \\ \bottomrule
\end{tabular}%
}
\vspace{-5pt}
\end{table} 
In the process of modifying and expanding the original queries, our primary goal was to introduce a diverse set of linguistic and structural variations that would challenge the model's ability to generalize beyond simple lookup-based responses.
Specifically, our construction process involves a systematic analysis of the original BIRD dataset, identifying queries of varying difficulty levels and involving different numbers of tables (1, 2, 3, and more than 3). We modify and expand these queries by altering table and column names, as well as filtering conditions, to create a more diverse set of challenges. In cases where the original dataset lacks sufficient examples with four or more tables (there are only 20 instances in BIRD-Dev dataset), queries that involved three tables are expanded to four. As shown in Table~\ref{tab:bigtable}, this process generates 50 new instances for each category, resulting in the ``BigTable'' dataset. Moreover, each item in the dataset underwent mutual verification by at least two individuals to ensure the accuracy.

\section{Evaluation}
\begin{figure}[htbp]  
  \begin{center}
    \includegraphics[width=0.8\textwidth]{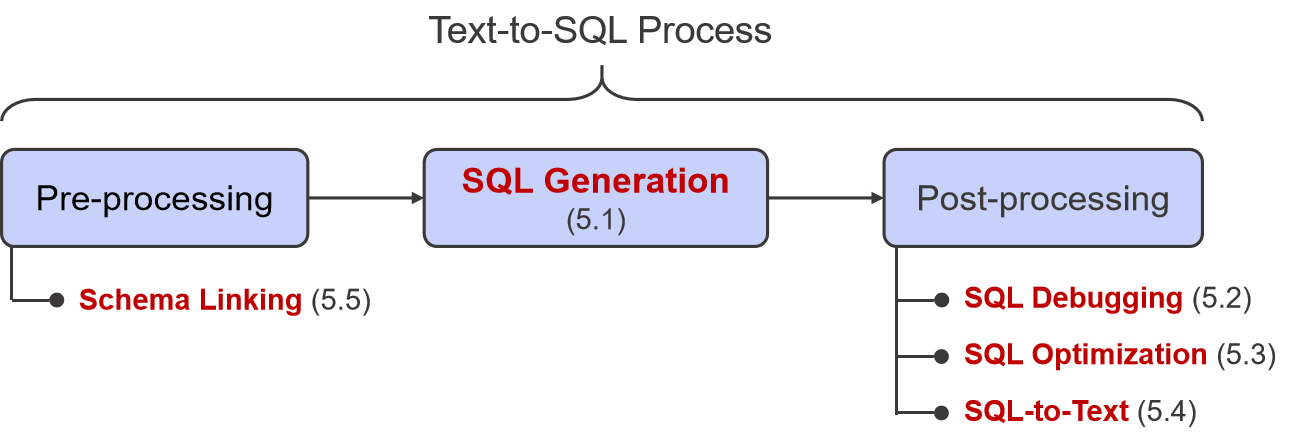} 
  \end{center}
  \vspace{-15pt}
  \caption{The paradigm for LLM-based Text-to-SQL and the tasks evaluated in SQLBench.}
    \label{fig:5tasks}
\end{figure}

In this section, we formally evaluate the different sub-tasks of SQLBench within the Text-to-SQL process to determine the performance differences among various LLMs and provide recommendations for addressing specific task requirements.
The relationships among the sub-tasks tested are illustrated in Figure~\ref{fig:5tasks}. The code is available at \url{https://anonymous.4open.science/r/SQLBench-E83A}.

\subsection{SQL Generation}\label{sec:text2sql}
\begin{table*}[ht]
\centering
\caption{EX (\%) of Text-to-SQL with various zero-shot prompt templates on Spider dev set.}
\label{table:zero-shot-prompting-optimization}
\setlength\tabcolsep{5pt}
{\footnotesize %
\begin{tabular}{@{}lcccc@{}}
\toprule
\textbf{Prompt Template} &
  \textbf{SQLCoder-34B} &
  \textbf{Codellama-34B} &
  \textbf{InternLM-70B} &
  \textbf{Llama2-Chat-70B} \\ \midrule
\textbf{DDL-HTML-Chat}         & 57.8 & 63.7          & \textbf{65.0} & 49.6                        \\
\textbf{DDL-HTML-Complete}     & 61.8 & \textbf{65.2} & 53.8 & 50.2                        \\
\textbf{DDL-MD-Chat}           & 63.2 & \textbf{68.4} & 66.3 & 48.7                        \\
\textbf{DDL-MD-Complete}       & 62.4 & \textbf{69.8} & 64.1 & 46.8                        \\
\textbf{DDL-Coding-Chat}       & 60.3 & \textbf{67.1} & 66.1 & 48.4                        \\
\textbf{DDL-Coding-Complete}   & 59.7 & \textbf{66.9} & 62.9 & 53.4                        \\
\textbf{SimpleDDL-MD-Chat} &
  \cellcolor[HTML]{EAF6F6}\textbf{65.0} &
  \cellcolor[HTML]{EAF6F6}\textbf{71.6} &
  \cellcolor[HTML]{EAF6F6}\textbf{67.4} &
  \cellcolor[HTML]{EAF6F6}\textbf{54.7} \\
\textbf{SimpleDDL-MD-Complete} & 63.3 & \textbf{66.0}  & 61.7 & 50.2 \\ \bottomrule
\end{tabular}%
}
\vspace{-5pt}
\end{table*}

Unlike previous learning-based studies, the primary challenge in LLM-based Text-to-SQL is the design of an effective prompt template $\mathcal{P}$ (as introduced in Section~\ref{sec:task-formulation}) for LLMs to generate accurate SQL queries, known as prompt engineering. Researchers have evaluated a variety of prompt templates~\cite{dail-sql}. However, these representations lack uniformity in their structure, making it difficult to find out how a specific feature within a prompt template impacts performance.

To address this issue, we investigate a more unified series of prompt templates. As shown in Listing~\ref{list:prompt} (a)-(c), these templates differ across three features:
\begin{itemize} \itemsep -1pt
    \item \textbf{DDL/SimpleDDL prefix} affects the representation of the database schema $\mathcal{S}$. 
         ``DDL'' (Data Definition Language) encompasses the standardized language that includes commands for defining the structure and properties of a database.
         ``SimpleDDL'' is simplified by only supplying table and column names.
    
    \item \textbf{MD/HTML/Coding infix} wraps the entire prompt template with Markdown syntax, HTML snippets and code comment blocks. 
    \item \textbf{Complete/Chat postfix} indicates the task of either completing SQL statements based on the "SELECT" clause or directly answering questions. 
\end{itemize}

\renewcommand{\figurename}{Listing}
\setcounter{figure}{0}
\begin{figure}[htbp]  
  \begin{center}
    \includegraphics[width=0.79\textwidth]{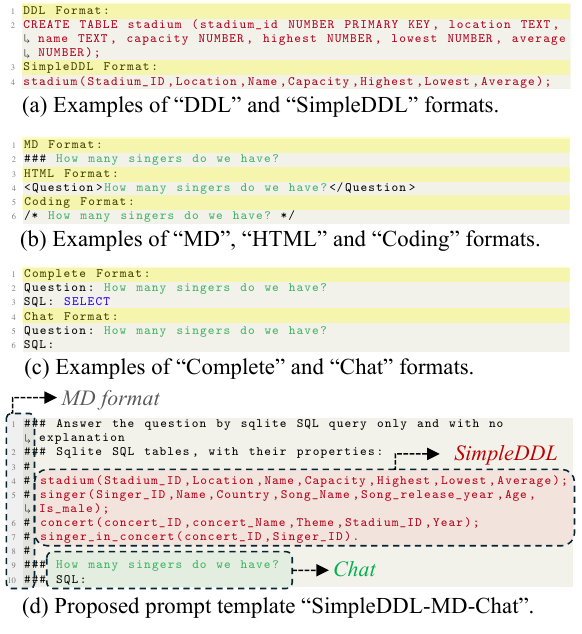} 
  \end{center}
  \vspace{-15pt}
  \caption{(a)-(c): different options for three format features; (d) the proposed optimal prompt templates.}
    \label{list:prompt}
\end{figure}

These features are combined to form a complete prompt template $\mathcal{P}$, and more details of these representations can be found in Appendix~\ref{app:text-to-sql prompt_template}.
We test these templates on Spider dev set. As shown in Table~\ref{table:zero-shot-prompting-optimization}, ``SimpleDDL-MD-Chat'' in Listing~\ref{list:prompt} (d) 
consistently outperforms all other prompts when applied to all 5 backbone LLMs.
%
To this end, we consistently utilize the prompt template ``SimpleDDL-MD-Chat'' throughout the subsequent evaluations in this paper. A detailed evaluation of ``BigTable'' can be found in Appendix~\ref{sec:bigtable_results}

\begin{conclusion}
    The prompt template ``SimpleDDL-MD-Chat'' achieves optimal performance in Text-to-SQL.
\end{conclusion}

\subsection{SQL Debugging}\label{sec:sql-debug}

In recent studies, researchers have demonstrated that LLMs possess self-reflection and self-correction capabilities similar to those of humans~\cite{react, self-reflexion, zhang2023controlling}.
Additionally, previous studies also investigate the potential of LLMs to debug the code they generate~\cite{self-debug,din-sql}. In this section, we provide a comprehensive analysis of the performance of numerous SQL debugging methods across LLMs.

\textbf{Debugging Dataset. }
We collect incorrect SQL queries generated by various LLMs in ``BigTable''
and visualize the distribution of their error information, as shown in Figure~\ref{fig:error word_cloud}. 
The error information can be divided into two categories:
(1) \textbf{System Error} refers to syntax errors in the SQL statement, and the detailed system error information is generated by the Database Management System (DBMS), e.g., ``syntax error'' and ``no such column'';
(2) \textbf{Result Error} indicates that the syntax of the SQL statement is correct, but the execution result does not match the ground truth. 

\renewcommand{\figurename}{Figure}
\setcounter{figure}{2}
\begin{figure*}[t]
\centering
\includegraphics[width=\textwidth]{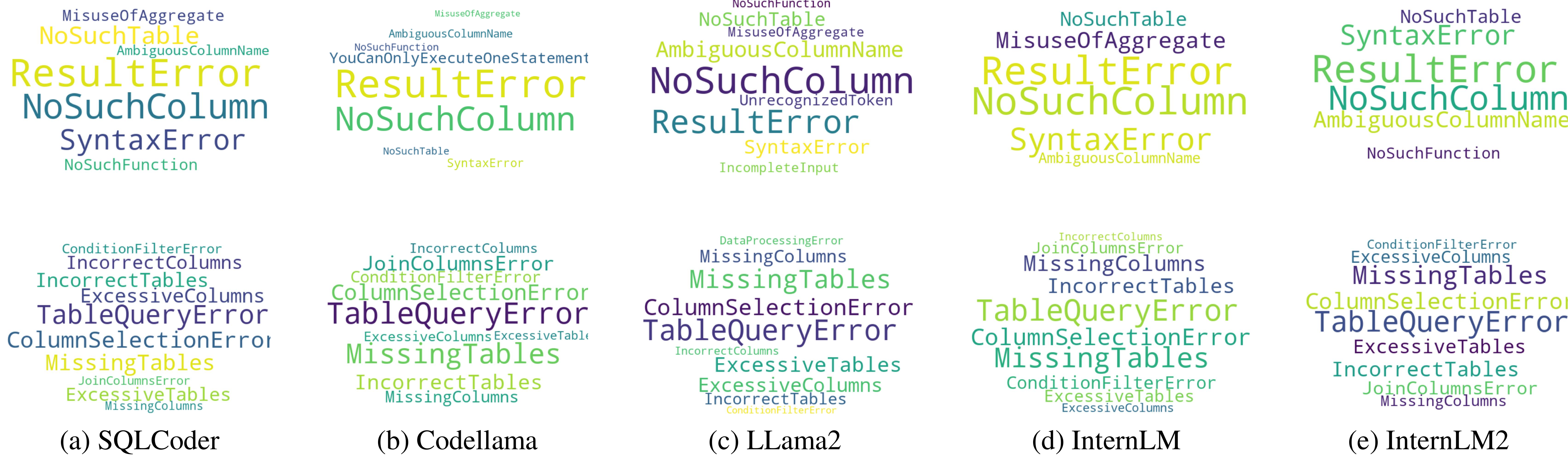}\vspace{-3mm}
\caption{Word cloud representation of error information for incorrect SQL queries generated by LLMs. \textbf{\emph{Top}}: System Error and Result Error. \textbf{\emph{Bottom}}: Detailed classification of Result Error.}
\label{fig:error word_cloud}
\vspace{-4mm}
\end{figure*}

\begin{figure}[htbp]  
  \begin{center}
    \includegraphics[width=0.79\textwidth]{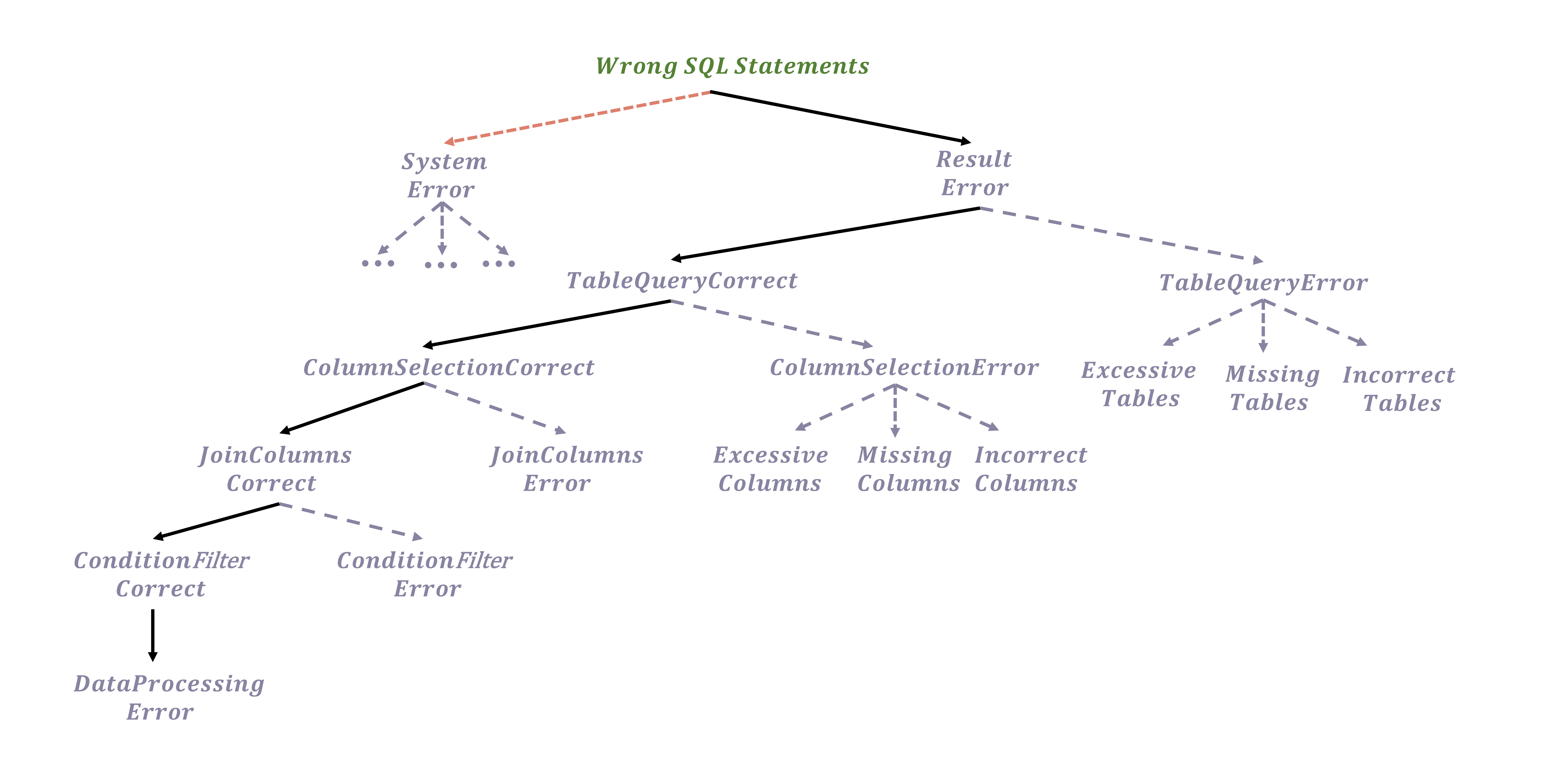} 
  \end{center}
  \vspace{-10pt}
  \caption{Process for determining error types. System Errors are directly got from the feedback of the SQL compiler. The five error types of Result Error are assessed sequentially based on their logical structure. When evaluating for Data Processing Errors, for example, it indicates that Table Query, Column Selection, and Join Column aspects have been correctly addressed.}
    \label{fig:error_cla}
    \vspace{-10pt}
\end{figure}

The word cloud distribution reveals that ``No Such Column'' and Result Error are the primary areas of error concentration for all models. 
However, the concise nature of the error information in the Result Error category significantly hampers the debugging performance. 
To this end, we categorize these errors based on the logical construction of SQL statements. As shown in Figure~\ref{fig:error_cla}, this classification is prioritized according to the logical structure within the SQL query. It is delineated in order into the following five subcategories:
(1) \textbf{Table Query Error} pertains to issues related to the selection of tables in the SQL query. It is further subdivided into three types: Excessive/Missing/Incorrect Tables, which respectively address scenarios where unnecessary tables are included, required tables are omitted, or the wrong tables are referenced;
(2) \textbf{Column Selection Error} focuses on the appropriateness of column selection. Similar to the Table Query Error, it is broken down into Excessive/Missing/Incorrect Columns;
(3) \textbf{Join Columns Error} examines the errors associated with JOIN operations; 
(4) \textbf{Condition Filter Error} encompasses errors that occur in the conditions used to filter the data, including incorrect comparisons or misapplied filters;
(5) \textbf{Data Processing Error} pertains to errors in the data processing stage, which includes aggregations, calculations, etc. applied to the data within the SQL query. 


\textbf{Debug Evaluation.}
To assess the impact of different levels of information granularity on performance, we propose 5 self-debugging strategies, progressively incorporating more details:
\begin{itemize}\itemsep -1pt
    \item \textbf{Regenerate}: Simply regenerate the SQL query with the same prompt in Section~\ref{sec:text2sql}. 
    \item \textbf{w/ Wrong SQL}: Let LLMs generate a new SQL query based on the wrong SQL statement.
    \item \textbf{w/ Wrong SQL + System\_error\_info}: Provide the wrong SQL statement, the corresponding System Error information and the rough Result Error information.
    \item \textbf{w/ Wrong SQL + All\_error\_info}: Add detailed Result Error information for those SQL queries that are syntactically correct  but semantically wrong. 
    \item \textbf{w/ Wrong SQL + All\_error\_info + Comment}: Add manual annotations for all error information. See Appendix~\ref{app:Debugging prompt_template} for a detailed prompt template.  
    
\end{itemize}

\begin{figure*}[t]
\centering
\includegraphics[width=\textwidth]{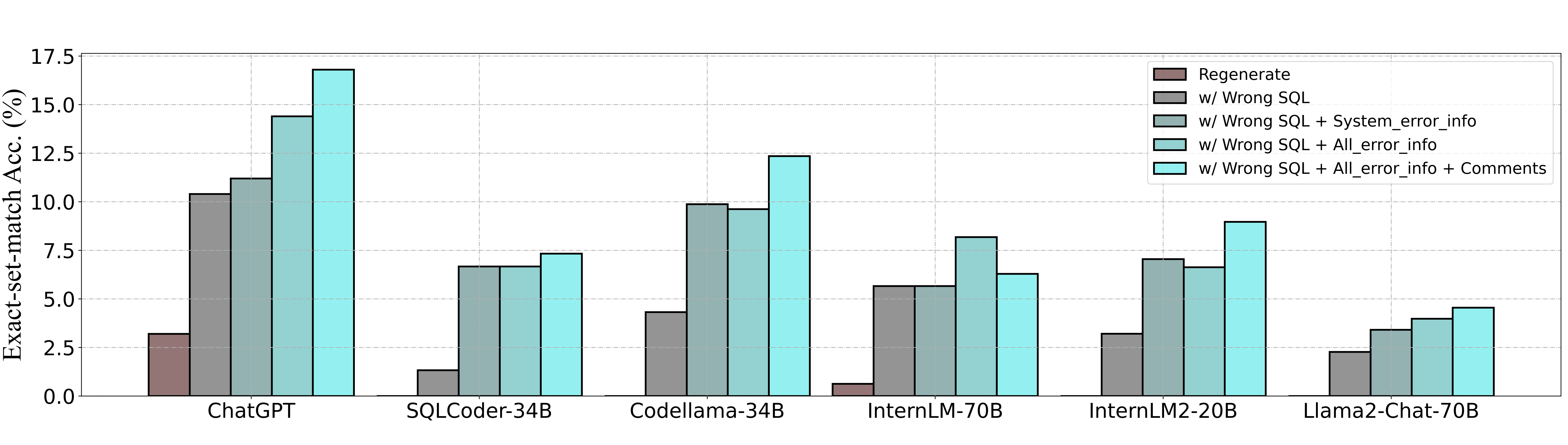}\vspace{-2mm}
\caption{EX (\%) improvement brought by self debug.}
\label{fig:self-debug}
\vspace{-4mm}
\end{figure*}

\textbf{\textit{What is the most powerful information organization of self debug?}}
As shown in Figure~\ref{fig:self-debug}, it is evident that the self-debugging performance of LLMs exhibits an upward trend with the introduction of more granular error information. 
In the absence of additional information, LLM does not possess the capability to regenerate correct answers. However, all models are able to comprehend fine-grained error information, whether it includes comments or not, and rectify their own mistakes.

\begin{conclusion}
     Detailed error information and corresponding annotations greatly enhance the capabilities of LLMs, enabling them to effectively correct errors.
\end{conclusion}

\begin{figure*}[t]
\centering
\includegraphics[width=\textwidth]{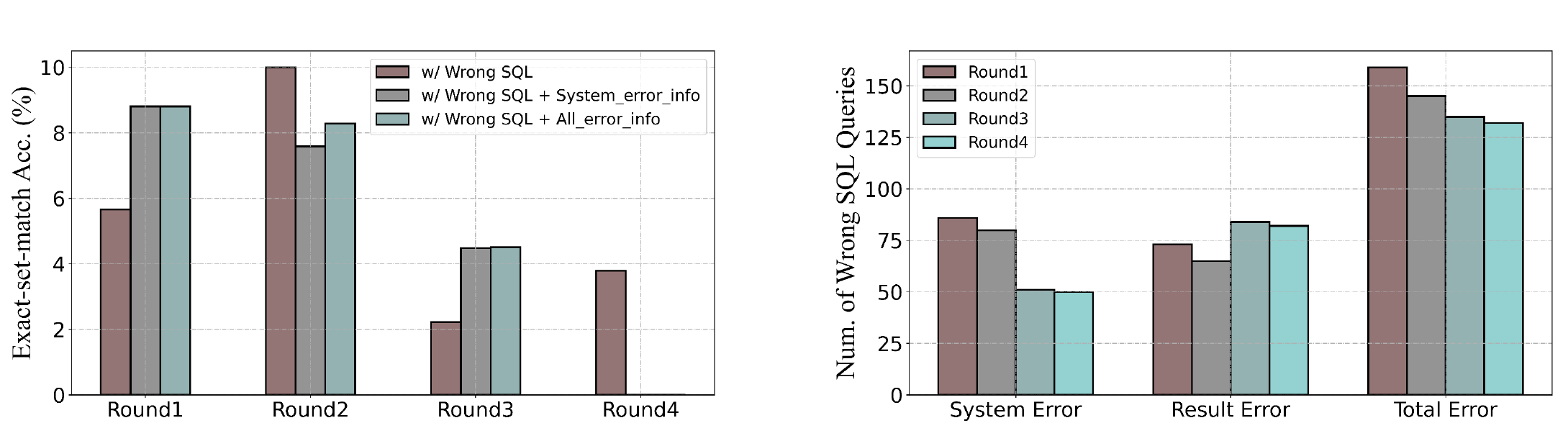}\vspace{-3mm}
\caption{\emph{\textbf{Left}}: EX (\%) improvement brought by multi-round self debug. \emph{\textbf{Right}}: Numbers of wrong SQL queries of detailed error type during the process of multi-round self debug.}
\label{fig:multi-round self-debug}
\vspace{-2mm}
\end{figure*}

\textbf{\textit{Can LLMs benefit from multi-round self debug?}} As shown in Figure~\ref{fig:multi-round self-debug}, substantial improvements in EX are achieved in the initial rounds of debugging, yet the performance gain become marginal later. This indicates that conducting 1-2 debugging rounds might strikes a favorable balance between performance improvement and economic efficiency. In addition, we analyze the distribution of detailed error types during the multi-round self-debugging process. As the debugging rounds advance, we observe a reduction in System Error and a slight rise in Result Error. This suggests that although syntax errors are fixed, the regenerated SQL statements may still have semantic errors. With each round of debugging, the generated statements tend to transition from System Error to Result Error before eventually converging towards correct SQL statements.

\begin{conclusion}
     Multi-round self-debugging aids in error correction for LLMs, but there exists a performance boundary, with 1-2 rounds of debugging being the optimal choice.
\end{conclusion}

\begin{table*}[htbp]
\centering
\caption{EX (\%) results of general debug. Each column demonstrates the debugging results of SQLCoder or InternLM for the wrong SQL statements generated by the corresponding model.}
\label{tab:general-debug}
\resizebox{\textwidth}{!}{%
\begin{tabular}{@{}llcccc@{}}
\toprule
\multicolumn{2}{l}{\textbf{Dataset $\longrightarrow$}} &
  \textbf{SQLCoder-34B} &
  \textbf{Codellama-34B} &
  \textbf{InternLM-70B} &
  \textbf{Llama2-Chat-70B} \\ \midrule
\multicolumn{6}{c}{\cellcolor[HTML]{E6E6E6}\textbf{SQLCoder Debugs All Errors}}                                                                                            \\
\multicolumn{2}{l}{\textbf{Regenerate}}                          & \cellcolor[HTML]{ECF4FF}\textbf{0.00} & 12.42 & 13.21                                 & 18.75 \\
\multicolumn{2}{l}{\textbf{w/ Wrong SQL}}                        & \cellcolor[HTML]{ECF4FF}\textbf{0.66} & 2.48  & 6.92                                   & 7.95  \\
\multicolumn{2}{l}{\textbf{w/ Wrong SQL + System\_error\_info}}  & \cellcolor[HTML]{ECF4FF}\textbf{4.64} & 11.18 & 9.43                                  & 11.36 \\
\multicolumn{2}{l}{\textbf{w/ Wrong SQL + All\_error\_info}} & \cellcolor[HTML]{ECF4FF}\textbf{4.64} & 11.80  & 8.81                                  & 10.80 \\
\multicolumn{6}{c}{\cellcolor[HTML]{E6E6E6}\textbf{InternLM Debugs All Errors}}                                                                                                \\
\multicolumn{2}{l}{\textbf{Regenerate}}                          & 9.27                                  & 11.80 & \cellcolor[HTML]{ECF4FF}\textbf{0.63} & 15.34 \\
\multicolumn{2}{l}{\textbf{w/ Wrong SQL}}                        & 7.95                                  & 10.56 & \cellcolor[HTML]{ECF4FF}\textbf{5.66} & 9.66  \\
\multicolumn{2}{l}{\textbf{w/ Wrong SQL + System\_error\_info}}  & 10.60                                 & 10.56 & \cellcolor[HTML]{ECF4FF}\textbf{8.81} & 11.93 \\
\multicolumn{2}{l}{\textbf{w/ Wrong SQL + All\_error\_info}} &
  11.92 &
  11.80 &
  \cellcolor[HTML]{ECF4FF}\textbf{8.81} &
  13.07 \\ \bottomrule
\end{tabular}%
}
\end{table*}

\textbf{\textit{Can an LLM debug the error incurred by other LLMs (general debugging)?}} 
As depicted in Table~\ref{tab:general-debug}, we select two representative LLMs, SQLCoder and InternLM, to assess their general debugging capabilities. 
When debugging erroneous SQL statements generated by other LLMs, additional error information can potentially impair the performance. 
Conversely, the most naive approach of simply regenerating the SQL statements often yields the best results. This highlights the differences between different LLMs.
The debugger LLMs may not encounter errors caused by other LLMs, and  these error type informations might confuse them. Consequently, it is unlikely to achieve performance improvement through general debugging. However, the integration of results generated by different LLMs holds promise as a future research direction.

\begin{conclusion}
    The performance of cross-LLM SQL debugging is inferior to the direct regeneration. A multi-agent approach that integrates outputs from different models shows great potential.
\end{conclusion}




\subsection{SQL Optimization}\label{sec:sql-optimization}

Execution efficiency of SQL queries is a critical aspect, particularly in real-time systems that utilize large-scale databases. 
In this section, we further explore whether LLMs are able to enhance the execution efficiency of correct SQL queries.

VES (see Equation~\ref{ves}) is commonly employed to evaluate the efficiency of SQL query execution. However, in practice, LLMs can sometimes rewrite a correct SQL query into an incorrect one, making it challenging to figure out if the main reason for the decline in VES is due to these incorrect rewrites or a decrease in the SQL execution efficiency. 
To this end, we suggest adopting a complementary metric C-VES (Correct-VES), which only assesses the execution efficiency of correct SQL queries: 
\begin{gather}
\tiny
{\rm C\text{-}VES}=\frac{\sum_{n\in\mathcal{N}_c} R(\mathcal{Y}_{n}^{\rm o},\hat{\mathcal{Y}}_n)}{|\mathcal{N}_c|}, \\
\quad \mathrm{where} \; R(\mathcal{Y}_{n}^{\rm o},\hat{\mathcal{Y}}_n)=\sqrt{\frac{\mathbf{E}(\mathcal{Y}_n^{\rm o})}{\mathbf{E}(\hat{\mathcal{Y}}_n)}}, \;
\mathcal{N}_c \doteq \{n | \mathbbm{1}(V_n,\hat{V}_n) = 1\}.
\end{gather}
Here, $\mathcal{Y}_n^{\rm o}$ and $\hat{\mathcal{Y}}_n$ denote the optimized predicted SQL and GT SQL for the $n$-th instance, respectively. $\mathbf{E}(\cdot)$ serves as a function that quantifies the absolute execution efficiency (e.g., the reciprocal of execution time) of each SQL within a specified environment. $\mathcal{N}_c$ represent the set of accurate SQLs (see Appendix~\ref{app:metrics} for detailed notations). C-VES is designed exclusively to validate the capability of LLMs to generate more efficient SQL queries, regardless of the potential drawback of rewriting correct SQLs into erroneous ones.

\textbf{\textit{Do LLMs have the capability for SQL self-optimization?}}
To the best of our knowledge, we are the first to consider utilizing LLMs for SQL optimization.
Specifically, we devise an extensive suite of prompts $\mathcal{P}_{\rm o}$ curated to SQL optimization:
\vspace{-5pt}
\begin{itemize}\itemsep -1pt
    \item \textbf{w/ $\mathcal{Y}$}: In this basic form, only original SQL statements are provided.
    \item \textbf{w/ $\mathcal{Y} + \mathcal{S} + \mathcal{Q}$}: Further incorporates the database schema $\mathcal{S}$ and the user question $\mathcal{Q}$.
    \item \textbf{w/ demo}: Introduce few-shot demonstrations without explanations. 
    \item \textbf{w/ demo + comments}: Add an explanation for the few-shot demonstrations. 
\end{itemize}

\begin{table*}[htbp]
\centering
\caption{VES and C-VES results of different SQL optimization methods.}
\label{tab:sql-optimization}
\resizebox{\textwidth}{!}{%
\begin{tabular}{@{}llllccccc@{}}
\toprule
\textbf{Methods} &
  \multicolumn{2}{l}{\textbf{Prompt Template}} &
  \textbf{Metrics} &
  \multicolumn{1}{l}{\textbf{ChatGPT}} &
  \multicolumn{1}{l}{\textbf{SQLCoder-34B}} &
  \multicolumn{1}{l}{\textbf{Codellama-34B}} &
  \multicolumn{1}{l}{\textbf{InternLM-70B}} &
  \multicolumn{1}{l}{\textbf{InternLM2-20B}} \\ \midrule
\multirow{2}{*}{\textbf{Baseline}} &
  \multicolumn{2}{l}{\multirow{2}{*}{\textbf{SimpleDDL-MD-Chat}}} &
  \textbf{VES} &
  36.90 &
  24.28 &
  18.94 &
  22.63 &
  19.81 \\
 &
  \multicolumn{2}{l}{} &
  \textbf{C-VES} &
  102.50 &
  101.17 &
  99.68 &
  110.39 &
  94.33 \\ \midrule
\multirow{10}{*}{\textbf{Two-Stage Generation}} &
  \multirow{6}{*}{\textbf{zero-shot}} &
  \multirow{2}{*}{\textbf{with $\mathcal{Y}$}} &
  \textbf{VES} &
  30.73 &
  16.86 &
  17.34 &
  15.33 &
  20.28 \\
 &
   &
   &
  \textbf{C-VES} &
  102.43 &
  102.19 &
  102.08 &
  102.17 &
  101.42 \\ \cmidrule(l){3-9} 
 &
   &
  \multirow{2}{*}{\textbf{w/ $\mathcal{Y} + \mathcal{S} + \mathcal{Q}$}} &
  \textbf{VES} &
  32.21 &
  18.90 &
  19.28 &
  18.44 &
  20.77 \\
 &
   &
   &
  \textbf{C-VES} &
  102.24 &
  102.14 &
  101.47 &
  102.43 &
  101.30 \\ \cmidrule(l){2-9} 
 &
  \multirow{4}{*}{\textbf{few-shot}} &
  \multirow{2}{*}{\textbf{w/ demo}} &
  \textbf{VES} &
  32.19 &
  18.84 &
  18.34 &
  17.97 &
  20.86 \\
 &
   &
   &
  \textbf{C-VES} &
  102.18 &
  101.84 &
  101.88 &
  102.68 &
  101.75 \\ \cmidrule(l){3-9} 
 &
   &
  \multirow{2}{*}{\textbf{w/ demo + comments}} &
  \textbf{VES} &
  32.65 &
  18.28 &
  18.52 &
  17.45 &
  20.92 \\
 &
   &
   &
  \textbf{C-VES} &
  102.03 &
  101.54 &
  102.86 &
  102.66 &
  102.06 \\ \midrule
\multirow{2}{*}{\textbf{Direct Generation}} &
  \multicolumn{2}{l}{\multirow{2}{*}{\textbf{SimpleDDL-MD-Chat-Efficiency}}} &
  \textbf{VES} &
  \textbf{39.26} &
  \textbf{27.77} &
  \textbf{20.75} &
  \textbf{25.23} &
  \textbf{25.93} \\
 &
  \multicolumn{2}{l}{} &
  \textbf{C-VES} &
  \textbf{103.31} &
  \textbf{102.84} &
  \textbf{103.75} &
  \textbf{102.98} &
  \textbf{101.70} \\ \bottomrule
\end{tabular}%
}
\end{table*}


These methods focus on the second stage optimization of the SQL queries that have been correctly generated by LLMs during the Text-to-SQL stage. 
In addition, we also evaluate a direct genaration method \textbf{SimpleDDL-MD-Chat-Efficiency}: Require LLMs to directly generate the most efficient SQL query statement based on user query.
The effectiveness of these SQL optimization methods is shown in Table~\ref{tab:sql-optimization}. 
Almost all two-stage methods experience a significant decrease in VES.
It can be attributed to the possibility of LLMs optimizing the correct SQL statements into incorrect ones, resulting in a further decrease in accuracy. Even when considering only the correct results, the performance improvement in terms of execution efficiency brought by the optimized SQL statements is almost negligible.

\begin{conclusion}
In-context learning methods present challenges in achieving effective SQL optimization with LLMs. 
\end{conclusion}

\subsection{SQL-to-Text}
\label{sec:sql2text}


The goal of SQL-to-Text is to transform the SQL query back into its original natural language question~\cite{xu2018sql2text, ma2021sql2text, 2021logic2text}.
It can assist researchers in evaluating the semantic comprehension capabilities of different LLMs, thus facilitating the development of more effective Text-to-SQL methodologies.

\begin{table*}[htbp]
\centering
\caption{SQL-to-Text performance of different LLMs, including the F1 scores of Rouge and BertScore, as well as the accuracy rate assessed by LLM.}
\label{fig:sql2text}
\resizebox{\textwidth}{!}{%
\begin{tabular}{@{}lcccccc@{}}
\toprule
{\color[HTML]{FF0000} \textbf{}} &
  \textbf{ChatGPT} &
  \textbf{SQLCoder-34B} &
  \textbf{Codellama-34B} &
  \textbf{InternLM-70B} &
  \textbf{InternLM2-20B} &
  \textbf{Llama2-Chat-70B} \\ \midrule
\textbf{Rouge-1}             &\cellcolor[HTML]{ECF4FF}\textbf{0.428}  & 0.118  & 0.359  & 0.408                                  & 0.410  & 0.368  \\
\textbf{Rouge-2}             &\cellcolor[HTML]{ECF4FF}\textbf{0.384}  & 0.039  & 0.167  & 0.198                                  & 0.294  & 0.159  \\
\textbf{Rouge-L}             &\cellcolor[HTML]{ECF4FF}\textbf{0.545}  & 0.108  & 0.320  & 0.372 & 0.358                                   & 0.325  \\
\textbf{BertScore}           &\cellcolor[HTML]{ECF4FF}\textbf{0.887} & 0.779  & 0.856  & 0.886                                  & \cellcolor[HTML]{ECF4FF}\textbf{0.887}  & 0.885  \\
\textbf{ChatGPT Evaluator} &\cellcolor[HTML]{ECF4FF}\textbf{70.5\%}  & 8.5\% & 47.0\% & 63.5\%                                 & 66.5\% & 50.0\% \\ 
\textbf{InternLM2 Evaluator} &62\%  & 22.5\% & 49.8\% & 65.8\%                                 & \cellcolor[HTML]{ECF4FF}\textbf{71.3\%} & 60.0\% \\ \bottomrule

\end{tabular}%
}
\end{table*}

We assess the performance of SQL-to-Text across different LLMs (See Appendix~\ref{app:sql2text} for prompt templates). 
The selected metrics for evaluation encompass the F1 values of Rouge-1/2/L and BertScore, along with the application of LLM to assess the semantic coherence between the two texts.
The evaluation results are depicted in Table~\ref{fig:sql2text}. ChatGPT and InternLM2 demonstrate the highest performance, followed by InternLM, while Codellama and SQLCoder exhibit comparatively lower performance. This highlights that even in regards to semantic description of code, general-purpose models exhibit significantly stronger descriptive capabilities compared to coding-specific models.
\begin{conclusion}
    Utilizing a general-purpose model for semantic description of SQL statements is a better choice.
\end{conclusion}

\subsection{Schema Linking}

Schema linking is recognized as a crucial prerequisite of generating correct SQL queries~\cite{yang2024sql}.
It involves aligning entity references in the question with the corresponding schema tables or columns, requiring the model to understand both structure and value of the database, as well as the the semantics of user questions. 
In LLM-based Text-to-SQL, prior studies~\cite{c3, din-sql} design prompt instructions with in-context learning examples to enable LLMs to retrieve linked tables and columns, which are then used for the downstream Text-to-SQL task.
However, there is currently no comprehensive evaluation of the performance of multiple diverse methods across various backbone LLMs.
%
In this section, we aim to bridge these gaps by: (1) introducing a elaborately designed metrics to assess schema linking methods; (2) presenting a novel schema linking method ``PreSQL'', which demonstrates superior performance; (3) conducting a comprehensive evaluation of a range of schema linking methods across various LLMs.

\begin{table*}[htbp]
\centering
\caption{RES results of different Schema Linking methods.}
\label{tab:schema-linking}
\resizebox{\textwidth}{!}{%
\begin{tabular}{@{}lcccccccccccc@{}}
\toprule
\textbf{} &
  \multicolumn{2}{c}{\textbf{ChatGPT}} &
  \multicolumn{2}{c}{\textbf{SQLCoder-34B}} &
  \multicolumn{2}{c}{\textbf{InternLM-70B}} &
  \multicolumn{2}{c}{\textbf{Codellama-34B}} &
  \multicolumn{2}{c}{\textbf{InternLM2-20B}} &
  \multicolumn{2}{c}{\textbf{Llama2-Chat-70B}} \\ \cmidrule(lr){2-3} \cmidrule(lr){4-5} \cmidrule(lr){6-7} \cmidrule(lr){8-9} \cmidrule(l){10-11} \cmidrule(l){12-13} 
 &
  w/o fk &
  w/ fk &
  w/o fk &
  w/ fk &
  w/o fk &
  w/ fk &
  w/o fk &
  w/ fk &
  w/o fk &
  w/ fk &
  w/o fk &
  w/ fk \\ \midrule
\textbf{Zero Shot} &
  0.6384 &
  0.6399 &
  0.2278 &
  0.4686 &
  0.5745 &
  0.5652 &
  0.3675 &
  0.4835 &
  0.5687 &
  0.5811 &
  0.4759 &
  0.5566 \\
\textbf{Few Shot} &
  0.6222 &
  0.6402 &
  0.3657 &
  0.3919 &
  0.5302 &
  0.4961 &
  0.4401 &
  0.4745 &
  0.464 &
  0.4829 &
  0.4123 &
  0.5375 \\
\textbf{PreSQL} &
  0.6888 &
  0.6610 &
  \cellcolor[HTML]{ECF4FF}\textbf{0.5661} &
  \cellcolor[HTML]{ECF4FF}\textbf{0.6417} &
  0.4632 &
  0.4881 &
  0.5085 &
  \cellcolor[HTML]{ECF4FF}\textbf{0.597} &
  0.5649 &
  0.6267 &
  0.4478 &
  0.5273 \\
\textbf{Few Shot + PreSQL} &
  \cellcolor[HTML]{ECF4FF}\textbf{0.7340} &
  \cellcolor[HTML]{ECF4FF}\textbf{0.7469} &
  0.5354 &
  0.6336 &
  \cellcolor[HTML]{ECF4FF}\textbf{0.5999} &
  \cellcolor[HTML]{ECF4FF}\textbf{0.6054} &
  \cellcolor[HTML]{ECF4FF}\textbf{0.5576} &
  0.5646 &
  \cellcolor[HTML]{ECF4FF}\textbf{0.6594} &
  \cellcolor[HTML]{ECF4FF}\textbf{0.7016} &
  \cellcolor[HTML]{ECF4FF}\textbf{0.6105} &
  \cellcolor[HTML]{ECF4FF}\textbf{0.6418} \\ \bottomrule
\end{tabular}%
}
\end{table*}
%
\textbf{\textit{Evaluation Metric for Schema Linking.}} 
%
%
The objective of schema linking is to retrieve all GT tables while avoiding the retrieval of excessive tables (with minimal redundancy).
To evaluate this, we design a comprehensive metric called \textbf{Retrieval Efficiency Score (RES)}:
\begin{gather}
\tiny
{\rm RES}=\frac{\sum_{n=1}^N\mathbbm{1}(T_n,\hat{T}_n)\cdot R(T_{n},\hat{T}_n)}N,
\\ \mathrm{where} \quad  \mathbbm{1}(T_n,\hat{T}_n)=\begin{cases}1,\ \mathrm{if} \; T_n\subseteq\hat{T}_n\\0,\ \mathrm{if} \; T_n\nsubseteq\hat{T}_n \end{cases}\!\!\!\!\!\!, \quad  R(T_{n},\hat{T_n})=\sqrt{\frac{|T_n|}{|\hat{T_n}|}},
\end{gather}
where $T_n$ and $\hat{T}_n$ denote the set of GT tables and retrieved tables for the $n$-th instance, respectively. $|\cdot|$ refers to the scale of a set. 




\textbf{Schema Linking Evaluation.}
We evaluate the most representative \textbf{Zero Shot} and \textbf{Few Shot} methods, which are proposed by C3~\cite{c3} and DIN-SQL~\cite{din-sql} respectively. 
In addition, a \textbf{PreSQL} method is introduced. First, we employ the zero-shot Text-to-SQL described in Section~\ref{sec:text2sql} to generate a preliminary SQL query. From this preliminary SQL, table and column entities are parsed to serve as the retrieval results.
Moreover, \textbf{Few Shot + PreSQL} takes the union of the retrieval results from both the Few Shot and PreSQL methods, aiming to leverage the strengths of each.

Note that we organize the information of the database schema in a way similar to ``SimpleDDL-'' in the prompt of all the methods mentioned above, which ignores the information about foreign keys. 
However, as argued in~\cite{rat-sql}, foreign keys embody features of known schema relations and provide important clues for understanding the database structure.
To this end, we conduct experiments under both settings, w/ and w/o foreign keys, to investigate how incorporating foreign keys in the prompt influences the performance of schema linking. Results are demonstrated in Table~\ref{tab:schema-linking} (refer to Appendix~\ref{app:exp_result} for results on detailed metrics like Exact Match \& Subset Match).

\textbf{\textit{Which method achieved the best performance in schema linking?}}
It can be seen that code-specific models excel in performance when utilizing the PreSQL approach, whereas general-purpose models yield optimal results through the Few-Shot + PreSQL method. This aligns with our expectations, as these two types of models excel in coding tasks (see Section~\ref{sec:text2sql}) and semantic understanding tasks (see Section~\ref{sec:sql2text}), respectively.

\textbf{\textit{Can foreign key information facilitate schema linking?}}
The introduction of foreign key information yield improved performance across all methods and all LLMs. This is evident since a valid JOIN operation in SQL queries is typically based on foreign keys. The foreign key information helps the model retrieve more ground truth tables by indicating all potential table pairs involved in a JOIN operation.

\begin{conclusion}
    Foreign key information is capable of advance the performance of schema linking. PreSQL yields the highest performance on coding-specific models, and integrating the results from Few Shot can further enhance performance on general-purpose models. 
\end{conclusion}

\section{Conclusion and Future Work}
\label{sec:conclusion}
In this study, we conduct a systematic benchmarking of the various sub-tasks within the Text-to-SQL pipeline, encompassing Text-to-SQL, SQL-Debugging, SQL-Optimization, SQL-to-Text, and Schema-Linking. 
Our comprehensive evaluation involves six distinct LLMs, spanning both general-purpose and coding-specific models.
We focus on determining the optimal prompt templates for each task, assessing performance variations among different approaches, and identifying the distinct capabilities and limitations of each LLM.
The results of the study demonstrate notable performance variations across the LLMs, underscoring the significance of careful model selection and prompt engineering in attaining optimal outcomes in text-to-SQL tasks.
Our benchmarking provides a meticulous perspective on the pipeline, equipping the research community with strategies to improve the semantic understanding and computational performance of LLMs. This advancement contributes to the development of more reliable Text-to-SQL systems.

As LLMs have seen rapid advancements, their performance continues to be quickly surpassed. It is therefore likely that even more powerful models could achieve superior results on our tasks. However, our work focuses on examining the general insights regarding LLM-based Text-to-SQL systems, such as identifying optimal prompt templates and characterizing the performance differences between code-specific and general-purpose models. These common findings provide valuable guidance for further improving the capabilities of this important class of language-based data management tools.

\medskip

{
\small
\bibliography{neurips_2024}
\bibliographystyle{plainnat}

}


\newpage
\appendix
\section{Supplemental Evaluation Results}
\label{app:exp_result}

\subsection{The Performance of GPT-4}
\begin{figure}[htbp]
    \centering
    \includegraphics[width=0.45\textwidth]{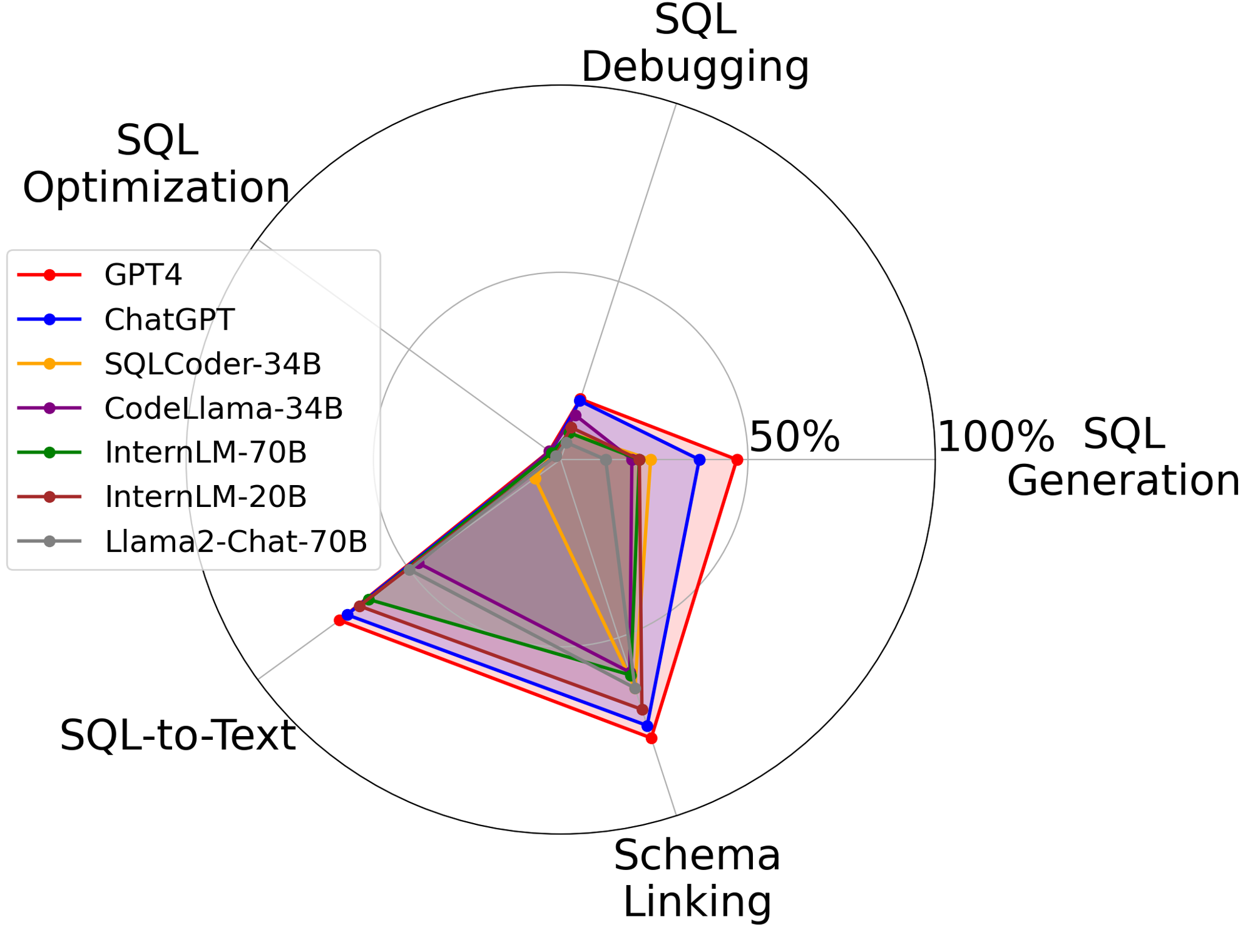}
    \caption{Overall performance of different LLMs in various sub-tasks of SQLBench. We add additional performance of GPT-4-turbo. GPT-4 consistently outperformed ChatGPT and other models across all sub-tasks. However, the performance gains are marginal. }
    \label{fig:gpt4}
\end{figure}

Based on the experiments (as shown in Figure~\ref{fig:gpt4}), GPT-4 consistently outperformed ChatGPT and other models across all sub-tasks. However, the performance gains were marginal, particularly in more challenging sub-tasks such as SQL optimization. These results underscore the difficulty of our benchmark and its significance in advancing the development of future Text-to-SQL models and approaches. Moreover, we believe that the core insights learned from our broader analysis will remain relevant even as specific model capabilities evolve.

\definecolor{eclipseBlue}{RGB}{42,0.0,255}
\definecolor{eclipseGreen}{RGB}{63,180,95}
\definecolor{eclipsePurple}{RGB}{175,0,25}
\definecolor{codewhite}{rgb}{0.70,0.70,0.70}
\definecolor{codegray}{rgb}{0.5,0.5,0.5}
\definecolor{codepurple}{rgb}{0.58,0,0.82}
\definecolor{backcolour}{rgb}{0.95,0.95,0.92}

\definecolor{instructioncolor}{rgb}{0.6275, 0.7686, 0.6157}
\definecolor{taskcolor}{rgb}{0.8824, 0.9255, 0.7843}
\definecolor{prefixcolor}{rgb}{0.9922, 1, 0.6824}


\lstdefinelanguage{Prompt}{
	backgroundcolor=\color{backcolour},   
	keywordstyle=\color{magenta},
	numberstyle=\tiny\color{codegray},
	basicstyle=\ttfamily\footnotesize,
	breakatwhitespace=false,         
	breaklines=true,   
    breakindent=-5pt,
	captionpos=b,                    
	keepspaces=true,                 
	numbers=left,                    
	numbersep=5pt,                  
	showspaces=false,                
	showstringspaces=false,
	showtabs=false,                  
	tabsize=4,
	escapeinside={`}{`},
	morecomment = [s][\color{eclipseGreen}\bfseries]{How}{?},
        morecomment = [l][\color{eclipseBlue}\bfseries]{SELECT},
        morecomment = [l][\color{eclipsePurple}\bfseries]{\$\{DATABASE_SCHEMA\}},
        morecomment = [s][\color{eclipsePurple}\bfseries]{CREATE}{;},
        morecomment = [l][\color{eclipsePurple}\bfseries]{Table},
        morecomment = [l][\color{eclipsePurple}\bfseries]{stadium},
        morecomment = [l][\color{eclipsePurple}\bfseries]{singer},
        morecomment = [l][\color{eclipsePurple}\bfseries]{concert},
        morecomment = [l][\color{eclipsePurple}\bfseries]{singer_in_concert},
        morecomment = [l][\color{codewhite}\bfseries]{\$\{TARGET_QUESTION\}},
    postbreak={
       \mbox{
           \lst@linebreakbgrd
           \rotatebox[y=0.7ex]{180}{\color{black}$\Lsh\,$}
       }
    },
}

\lstdefinelanguage{Special_Prompt}{
	backgroundcolor=\color{backcolour},   
	keywordstyle=\color{magenta},
	numberstyle=\tiny\color{codegray},
	basicstyle=\ttfamily\footnotesize,
	breakatwhitespace=false,         
	breaklines=true,   
    breakindent=-5pt,
	captionpos=b,                    
	keepspaces=true,                 
	numbers=left,                    
	numbersep=5pt,                  
	showspaces=false,                
	showstringspaces=false,
	showtabs=false,                  
	tabsize=4,
	escapeinside={`}{`},
        moredelim=**[is][\color{eclipsePurple}\bfseries]{\&\&}{\&\&},
        moredelim=**[is][\color{eclipseGreen}\bfseries]{\^\^}{\^\^},
        moredelim=**[is][\color{eclipseBlue}\bfseries]{@@}{@@}, 
    postbreak={
       \mbox{
           \lst@linebreakbgrd
           \rotatebox[y=0.7ex]{180}{\color{black}$\Lsh\,$}
       }
    },
}

\lstdefinelanguage{Question}{
	backgroundcolor=\color{backcolour},   
	sensitive = true,
	morecomment = [s]{Gold}{SQL},
	commentstyle ={\color{red}\bfseries\underbar},
	morestring = [b]",
	morestring = [b]',
	stringstyle = \color{eclipseGreen},
	basicstyle=\ttfamily\footnotesize,
	breaklines=true,
	alsoletter=!?-,
	emph={StableVicuna-13B, Vicuna-13B, Gold-SQL, Write a sql to answer},
	emphstyle={\color{red}\bfseries\underbar},
	captionpos=b,
	escapeinside={[}{]},
	keepspaces=true,              
	showspaces=false,                
	showstringspaces=false,
	showtabs=false,                  
	tabsize=4,
	columns=flexible
}

\subsection{Additional Experimental Results on ``BigTable''}

\subsubsection{End-to-End Text-to-SQL evaluation.}
\label{sec:bigtable_results}

\begin{table*}[htbp]
\centering
\caption{EX (\%) of different LLMs on ``BigTable''. Prompt template is ``SimpleDDL-MD-Chat".}
\label{tab:text2sql-on-bigtable}
\resizebox{0.8\textwidth}{!}{%
\begin{tabular}{@{}lcccccc@{}}
\toprule
\textbf{No. of GT Tables} & \textbf{ChatGPT} & \textbf{SQLCoder-34B} & \textbf{Codellama-34B} & \textbf{InternLM-70B} & \textbf{Llama2-Chat-70B} & \textbf{InternLM2-20B} \\ \midrule
\textbf{1}               & \textbf{60.00} & 44.00 & 32.00 & 44.00 & 22.00 & 40.00 \\
\textbf{2}               & \textbf{20.00} & 8.00  & 10.00 & 10.00 & 10.00 & 6.00  \\
\textbf{3}               & \textbf{38.00} & 24.00 & 22.00 & 20.00 & 8.00  & 18.00 \\
\textbf{\textgreater{}3} & \textbf{30.00} & 20.00 & 12.00 & 10.00 & 8.00  & 20.00 \\
\textbf{Total}           & \textbf{37.00} & 24.00 & 19.00 & 21.00 & 12.00 & 21.00 \\ \bottomrule
\end{tabular}%
}
\end{table*}

This research conducts an end-to-end evaluation of the Text-to-SQL capabilities of various LLMs using the "SimpleDDL-MD-Chat" prompt template on the "BigTable" dataset, with results depicted in Table~\ref{tab:text2sql-on-bigtable}.

\textbf{Comparison of Performance Across Different Models.} The results demonstrate a clear performance hierarchy among the models, with SQLCoder, CodeLlama, InternLM, and InternLM2 consistently outperforming Llama2-Chat. This finding highlights the effectiveness of coding-specific models, such as SQLCoder and CodeLlama, in the Text-to-SQL domain. Additionally, certain general-purpose models, like InternLM and InternLM2, can achieve performance levels comparable to specialized models, even without fine-tuning for coding tasks.

\textbf{Difficulty Comparison Across Different Numbers of GT Tables.}
We examine the query difficulty based on the number of GT tables involved. The results reveal a decrease in EX as the number of GT tables increases. Notably, the EX for models on queries with two GT tables is unexpectedly lower compared to those with three or more GT tables. This observation can be attributed to the fact that queries involving two GT tables have the highest average number of columns (see Table~\ref{tab:bigtable}). 


\subsubsection{Schema Linking.}
For a more detailed analysis of schema linking, we introduce two supplemental metrics, Subset Match and Exact Match. These metrics measure the proportion of instances in which the GT tables are a subset of, or exactly match, retrieved tables, respectively. 
We observe from Table~\ref{tab:schema-linking_app} that the Few Shot + PreSQL and PreSQL method achieves the highest Subset Match and the Exact Match, respectively, which explains the higher RES achieved by these methods.

\begin{table*}[htbp]
\centering
\caption{The results of table linking include the metrics of exact match and subset match.}
\label{tab:schema-linking_app}
\resizebox{\textwidth}{!}{%
\begin{tabular}{@{}llllllllllll@{}}
\toprule
\textbf{} &
   &
  \multicolumn{2}{c}{\textbf{SQLCoder-34B}} &
  \multicolumn{2}{c}{\textbf{InternLM-70B}} &
  \multicolumn{2}{c}{\textbf{Codellama-34B}} &
  \multicolumn{2}{c}{\textbf{InternLM2-20B}} &
  \multicolumn{2}{c}{\textbf{Llama2-Chat-70B}} \\ \cmidrule(lr){3-4} \cmidrule(lr){5-6} \cmidrule(lr){7-8} \cmidrule(lr){9-10} \cmidrule(l){11-12} 
 &
   &
  \multicolumn{1}{c}{w/o fk} &
  \multicolumn{1}{c}{w/ fk} &
  \multicolumn{1}{c}{w/o fk} &
  \multicolumn{1}{c}{w/ fk} &
  \multicolumn{1}{c}{w/o fk} &
  \multicolumn{1}{c}{w/ fk} &
  \multicolumn{1}{c}{w/o fk} &
  \multicolumn{1}{c}{w/ fk} &
  \multicolumn{1}{c}{w/o fk} &
  \multicolumn{1}{c}{w/ fk} \\ \midrule
 &
  \textbf{Subset Match} &
  0.28 &
  0.55 &
  0.66 &
  0.645 &
  0.495 &
  0.655 &
  0.605 &
  0.67 &
  0.655 &
  0.76 \\
\multirow{-2}{*}{\textbf{Zero Shot}} &
  \textbf{Exact Match} &
  0.135 &
  0.295 &
  0.355 &
  0.34 &
  0.19 &
  0.27 &
  0.235 &
  0.415 &
  0.12 &
  0.135 \\ \midrule
 &
  \textbf{Subset Match} &
  0.5 &
  0.475 &
  0.595 &
  0.555 &
  0.645 &
  0.7 &
  0.53 &
  0.545 &
  0.57 &
  0.72 \\
\multirow{-2}{*}{\textbf{Few Shot}} &
  \textbf{Exact Match} &
  0.155 &
  0.22 &
  0.355 &
  0.34 &
  0.19 &
  0.155 &
  0.33 &
  0.355 &
  0.125 &
  0.18 \\ \midrule
 &
  \textbf{Subset Match} &
  0.6 &
  0.675 &
  0.475 &
  0.505 &
  0.525 &
  0.62 &
  0.6 &
  0.67 &
  0.51 &
  0.65 \\
\multirow{-2}{*}{\textbf{PreSQL}} &
  \cellcolor[HTML]{EFEFEF}\textbf{Exact Match} &
  \cellcolor[HTML]{EFEFEF}\textbf{0.465} &
  \cellcolor[HTML]{EFEFEF}\textbf{0.55} &
  \cellcolor[HTML]{EFEFEF}\textbf{0.43} &
  \cellcolor[HTML]{EFEFEF}\textbf{0.435} &
  \cellcolor[HTML]{EFEFEF}\textbf{0.45} &
  \cellcolor[HTML]{EFEFEF}\textbf{0.525} &
  \cellcolor[HTML]{EFEFEF}\textbf{0.5} &
  \cellcolor[HTML]{EFEFEF}\textbf{0.525} &
  \cellcolor[HTML]{EFEFEF}\textbf{0.3} &
  \cellcolor[HTML]{EFEFEF}\textbf{0.23} \\ \midrule
 &
  \cellcolor[HTML]{ECF4FF}\textbf{Subset Match} &
  \cellcolor[HTML]{ECF4FF}\textbf{0.715} &
  \cellcolor[HTML]{ECF4FF}\textbf{0.77} &
  \cellcolor[HTML]{ECF4FF}\textbf{0.67} &
  \cellcolor[HTML]{ECF4FF}\textbf{0.675} &
  \cellcolor[HTML]{ECF4FF}\textbf{0.755} &
  \cellcolor[HTML]{ECF4FF}\textbf{0.785} &
  \cellcolor[HTML]{ECF4FF}\textbf{0.74} &
  \cellcolor[HTML]{ECF4FF}\textbf{0.78} &
  \cellcolor[HTML]{ECF4FF}\textbf{0.815} &
  \cellcolor[HTML]{ECF4FF}\textbf{0.86} \\
\multirow{-2}{*}{\textbf{Few Shot + PreSQL}} &
  \textbf{Exact Match} &
  0.195 &
  0.345 &
  0.395 &
  0.405 &
  0.2 &
  0.185 &
  0.42 &
  0.46 &
  0.19 &
  0.165 \\ \bottomrule
\end{tabular}%
}
\end{table*}

\subsection{Additional Experimental Results on BIRD}

\begin{figure*}[t]
\centering
\includegraphics[width=0.8\textwidth]{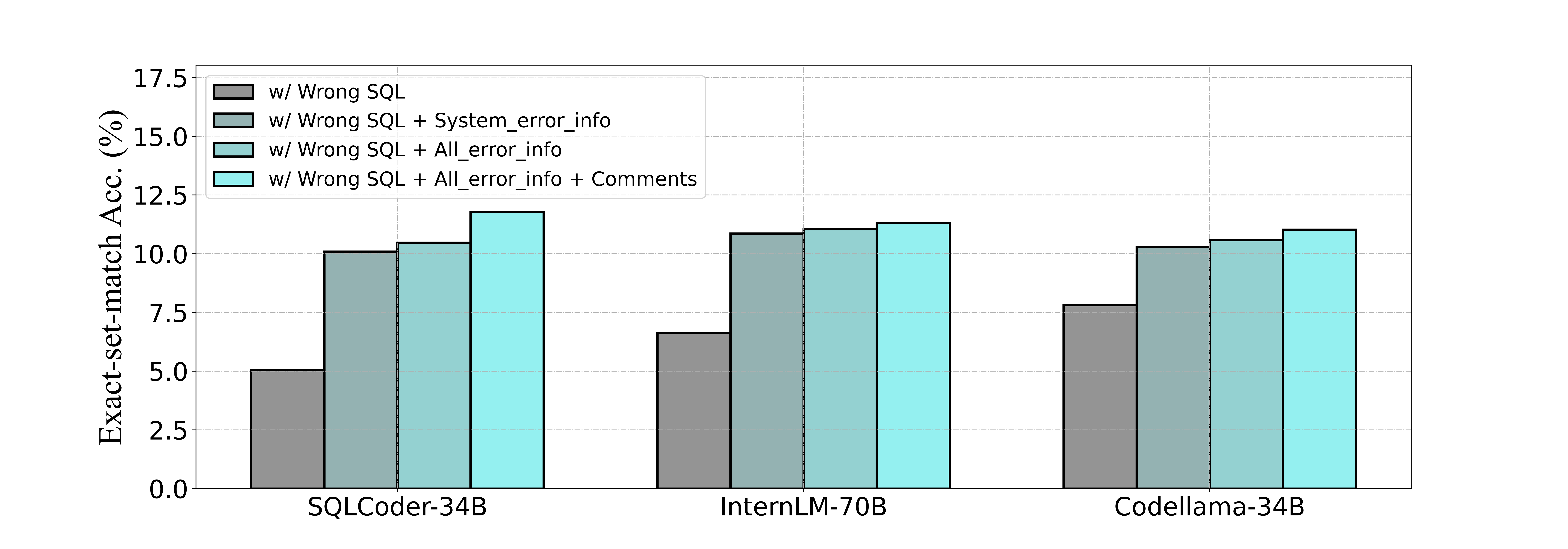}\vspace{-7mm}
\caption{EX (\%) improvement brought by self debug on BIRD dev set.}
\label{fig:self-debug-bird}
\end{figure*}

To further justify the effectiveness and robustness of the core conclusions of this paper, we conduct identical experiments on the BIRD development dataset. The BIRD dev set contains 1,534 instances, which exhibit a larger scale compared to ``BigTable''. In the experiments, we only choose representative LLMs and methods that are influential in deriving core conclusions. Results for five sub-tasks are shown in Table~\ref{tab:text2sql-bird}, Figure~\ref{fig:self-debug-bird}, Tables 11-13, respectively. These experimental results are largely consistent with the core conclusions obtained on ``BigTable'', supporting the validity of the conclusions.

\begin{table*}[htbp]
\centering
\footnotesize
\caption{EX (\%) of different LLMs on BIRD dev set. prompt template is ``SimpleDDL-MD-Chat".}
\label{tab:text2sql-bird}
\resizebox{0.7\textwidth}{!}{%
\begin{tabular}{@{}lcccc@{}}
\toprule
\textbf{No. of GT Tables} & \textbf{SQLCoder-34B} & \textbf{InternLM-70B} & \textbf{Codellama-34B} & \textbf{InternLM2-20B} \\ \midrule
\textbf{1}               & 40.33 & \textbf{43.37} & 40.88 & 40.33  \\
\textbf{2}               & \textbf{28.42}  & 25.24 & 27.57 & 28.31 \\
\textbf{3}               & \textbf{21.05} & 15.79 & 16.75  & \textbf{21.05} \\
\textbf{\textgreater{}3} & \textbf{30.00} & 5.00 & 15.00  & \textbf{30.00} \\
\textbf{Total}           & \textbf{30.25} & 27.97 & 29.07 & 30.18  \\ \bottomrule
\end{tabular}%
}
\end{table*}

\begin{table*}[htbp]
\centering
\footnotesize
\caption{VES results of SQL Optimization on BIRD dev set.}
\label{tab:sql-opt-bird}
\resizebox{0.7\textwidth}{!}{%
\begin{tabular}{@{}lccc@{}}
\toprule
\textbf{Prompt Template} & \textbf{SQLCoder-34B} & \textbf{InternLM-70B} & \textbf{Codellama-34B} \\ \midrule
with $\mathcal{Y}$               & 99.68 & 99.99 & 99.04   \\
w/ $\mathcal{Y} +\mathcal{S}$               & 102.65  & 100.10 & 101.48 \\
w/ $\mathcal{Y} + \mathcal{S} + \mathcal{Q}$           & 102.92 & 100.69 & 101.78  \\ 
SimpleDDL-MD-Chat-Efficiency & 101.26 & 100.51 & 100.34 \\ \bottomrule
\end{tabular}%
}
\end{table*}

\begin{table*}[H]
\centering
\caption{SQL-to-Text performance on BIRD dev set, including the F1 scores of Rouge and BertScore, as well as the accuracy rate assessed by LLM. }
\label{tab:sql2text-bird}
\resizebox{0.8\textwidth}{!}{%
\begin{tabular}{@{}lccc@{}}
\toprule
{\color[HTML]{FF0000} \textbf{}} &
  \textbf{Codellama-34B} &
  \textbf{InternLM-70B} &
  \textbf{Llama2-Chat-70B} \\ \midrule
\textbf{Rouge-1}             & 0.423  & \cellcolor[HTML]{ECF4FF}\textbf{0.495}                                   & 0.454   \\
\textbf{Rouge-2}             & 0.231  & \cellcolor[HTML]{ECF4FF}\textbf{0.273}                                    & 0.230    \\
\textbf{Rouge-L}             & 0.423  & \cellcolor[HTML]{ECF4FF}\textbf{0.449}   & 0.408                                    \\
\textbf{BertScore}           & 0.908  & \cellcolor[HTML]{ECF4FF}\textbf{0.924}                                  & 0.919    \\
\textbf{LLM Evaluator} & 64.0\% & \cellcolor[HTML]{ECF4FF}\textbf{80.8}\%                               & 75.3\%  \\ \bottomrule

\end{tabular}%
}
\end{table*}

\begin{table}[htbp]
\centering
\caption{RES results of Schema Linking on BIRD dev set.}
\label{tab:schema-linking-bird}
\resizebox{\columnwidth}{!}{%
\begin{tabular}{@{}lcccccc@{}}
\toprule
\textbf{} & \multicolumn{2}{c}{\textbf{SQLCoder 34B}} & \multicolumn{2}{c}{\textbf{InternLM-70B}} & \multicolumn{2}{c}{\textbf{Codellama-34B}} \\ \cmidrule(lr){2-3} \cmidrule(lr){4-5} \cmidrule(l){6-7} 
                           & w/o fk & w/ fk & w/o fk & w/ fk & w/o fk & w/ fk \\ \midrule
\textbf{Zero Shot} &
  0.7034 &
  0.7138 &
  0.7505 &
  0.7739 &
  \cellcolor[HTML]{ECF4FF}\textbf{0.7130} &
  0.6564 
 \\
\textbf{Few Shot} &
  0.4343 &
  0.4078 &
  0.7365 &
  0.7574 &
  0.5712 &
  0.5822 
 \\
\textbf{PreSQL} &
  \cellcolor[HTML]{ECF4FF}\textbf{0.7274} &
  \cellcolor[HTML]{ECF4FF}\textbf{0.7715} &
  0.5563 &
  0.5958 &
  0.6389 &
  \cellcolor[HTML]{ECF4FF}\textbf{0.6615} 
 \\
\textbf{Few Shot + PreSQL} &
  0.6621 &
  0.7296 &
  \cellcolor[HTML]{ECF4FF}\textbf{0.7686} &
  \cellcolor[HTML]{ECF4FF}\textbf{0.7936} &
  0.6551 &
  0.6417 
 \\ \bottomrule
\end{tabular}%
}
\end{table}

\section{Details for Dataset Construction}

In the process of modifying and expanding the original queries, our primary goal is to introduce a diverse set of linguistic and structural variations that would challenge the model's ability to generalize beyond simple lookup-based responses.

Specifically, we categorize the SQL statements into logical components, including Table Queries, Column Selection, Join Columns, Condition Filters, and Data Processing. This categorization is consistent with the approach we used in constructing the dataset for the SQL Debugging task. By systematically modifying these five components, we are able to generate new queries $\mathcal{Q}$ and SQL statements $\mathcal{Y}$ that differed from the original examples in Bird.
Furthermore, to ensure a balanced distribution of the number of tables involved in the queries, we expand queries with three tables to include four tables in cases where the original dataset lacked sufficient examples with four or more tables (as there are only 20 such instances in the BIRD-Dev dataset).

Listing~\ref{lst:example_construction} provides an example of expanding table queries and column selection by adding condition filters.

\begin{lstlisting}[language=Special_Prompt, label={lst:example_construction}, caption={An example of the construction of a sample in BigTable.}, float=htbp]
Original Question: ^^For the client who applied 98832 USD loan in 1996/1/3, when was his/her birthday?^^
Original  SQL: @@SELECT T3.birth_date FROM loan AS T1 INNER JOIN account AS T2 ON T1.account_id = T2.account_id INNER JOIN client AS T3 ON T2.district_id = T3.district_id WHERE T1.date = '1996-01-03' AND T1.amount = 98832@@
New Question:  ^^When is the birthday of a client who applied 98832 USD loan in East Bohemia in 1996/1/3? ^^
New SQL: @@SELECT T3.birth_date FROM loan AS T1 INNER JOIN account AS T2 ON T1.account_id = T2.account_id INNER JOIN client AS T3 ON T2.district_id = T3.district_id WHERE T1.date = '1996-01-03' AND T1.amount = 98832 AND T3.district_id = (SELECT district_id FROM district WHERE A3 = 'east Bohemia' )@@
\end{lstlisting}

\section{Details for Sub-tasks}
\subsection{Task Formulation}

The formulations for each sub-task are as follows:
\begin{itemize}
    \item SQL Debugging: Transforming the initial erroneous SQL queries $\mathcal{Y}^{\rm w}$ into the correct form $\mathcal{Y}^{\rm r}$: $\mathcal{Y}^{\rm r} = f_\mathcal{M}(\mathcal{P}_{\rm d}(\mathcal{Y}^{\rm w}, \mathcal{I}))$.
    \item SQL Optimization: Transforming the initial SQL queries $\mathcal{Y}$ into an optimized form $\mathcal{Y}^{\rm o}$, aiming to improve efficiency while preserving correctness: $\mathcal{Y}^{\rm o} = f_\mathcal{M}(\mathcal{P}_{\rm o}(\mathcal{Y}, \mathcal{I}))$.
    \item SQL-to-Text: Converting SQL queries $\mathcal{Y}$ into a natural language description $\mathcal{Q’}$ that effectively conveys the content of the query: $\mathcal{Q’} = f_\mathcal{M}(\mathcal{P}_{\rm t}(\mathcal{Y}, \mathcal{I}))$.
    \item Schema Linking: Generating a linking representation $\mathcal{L} = {(e_i, t_j, c_k)}$ based on the natural language query $\mathcal{Q}$ and the database schema $\mathcal{S}$ that maps entities or keywords $e_i$ in $\mathcal{Q}$ to tables $t_j$ and columns $c_k$ in schema $\mathcal{S}$: $\mathcal{L} = f_\mathcal{M}(\mathcal{P}_{\rm l}(\mathcal{Q}, \mathcal{S}))$.
\end{itemize}
Here, $\mathcal{P}$ and $\mathcal{I}$ are the corresponding prompt template and the additional information used for different tasks, $f_\mathcal{M}(\cdot)$ represents the mapping function of the LLM $\mathcal{M}$.

\subsection{Details for Evaluation Metrics}\label{app:metrics}
In this section, we present the detailed definition and discussion of evaluation metrics in this paper.

Two primary evaluation metrics for assessing the accuracy of SQLs on Spider are Exact Matching(EM) and \textbf{Execution Accuracy(EX)}. EM measures whether the predicted query as a whole is equivalent to the gold query. It is possible to encounter false negative evaluations since a question might be solvable by multiple syntactically different but semantically identical SQL statements. EX is a more widely used metric, measures whether the result of executing the predicted query matches the gold value. We use EX to evaluate the accuracy of SQLs in this paper, which is defined as:
\begin{gather}
{\rm EX}=\frac{\sum_{n=1}^N\mathbbm{1}(V_n,\hat{V}_n)}N, \\
\mathrm{where} \quad  \mathbbm{1}(V_n,\hat{V}_n)=\begin{cases}1,\ \mathrm{if} \; V_n = \hat{V}_n\\0,\ \mathrm{if} \; V_n \neq \hat{V}_n \end{cases}\!\!\!\!\!\!,
\end{gather}
where $V_n$ and $\hat{V}_n$ denote execution results of the predicted SQL and the GT SQL for the $n$-th instance in the evaluation set, respectively. 

BIRD further proposes \textbf{Valid Efficiency Score (VES)}, an integrated metric assessing both accuracy of execution results (i.e., EX) and the execution efficiency of SQL queries. VES is defined as:

\begin{gather}
{\rm VES}=\frac{\sum_{n=1}^N\mathbbm{1}(V_n,\hat{V}_n)\cdot R(\mathcal{Y}_{n}^{\rm o},\hat{\mathcal{Y}}_n)}N, 
\label{ves}\\
\mathrm{where} \quad R(\mathcal{Y}_{n}^{\rm o},\hat{\mathcal{Y}}_n)=\sqrt{\frac{\mathbf{E}(\mathcal{Y}_n^{\rm o})}{\mathbf{E}(\hat{\mathcal{Y}}_n)}},
\end{gather}
where $\mathcal{Y}_n^{\rm o}$ and $\hat{\mathcal{Y}}_n$ denote the optimized predicted SQL and GT SQL for the $n$-th instance, respectively. $\mathbf{E}(\cdot)$ serves as a function that quantifies the absolute execution efficiency (e.g., the reciprocal of execution time) of each SQL within a specified environment.

\subsection{Details for SQL Debugging}

\begin{figure}[htbp]
    \centering
    \includegraphics[width=0.5\textwidth]{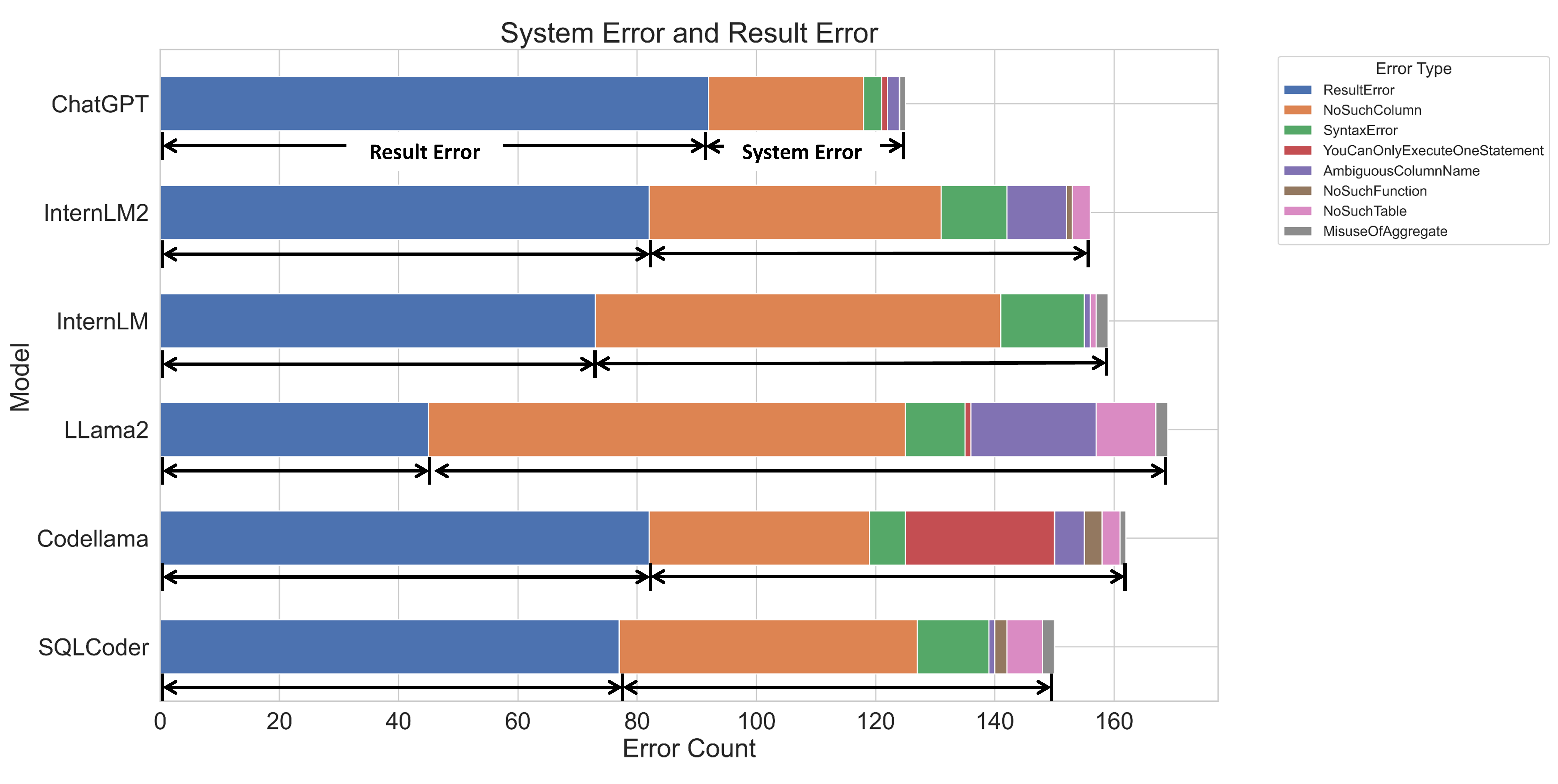}
    \vspace{-8pt}
    \caption{Error information of incorrect SQL statements generated by LLMs. Models with higher performance are less likely to commit basic System Errors (e.g., SQL statements that fail to execute and get errors information from the SQL compiler) and are more prone to Result Errors (i.e., SQL statements that execute correctly but yield incorrect answers). Therefore, it is necessary to further classify Result Errors.}
    \label{fig:all_error}
    \vspace{-10pt}
\end{figure}

\begin{figure}[htbp]
    \centering
    \includegraphics[width=0.5\textwidth]{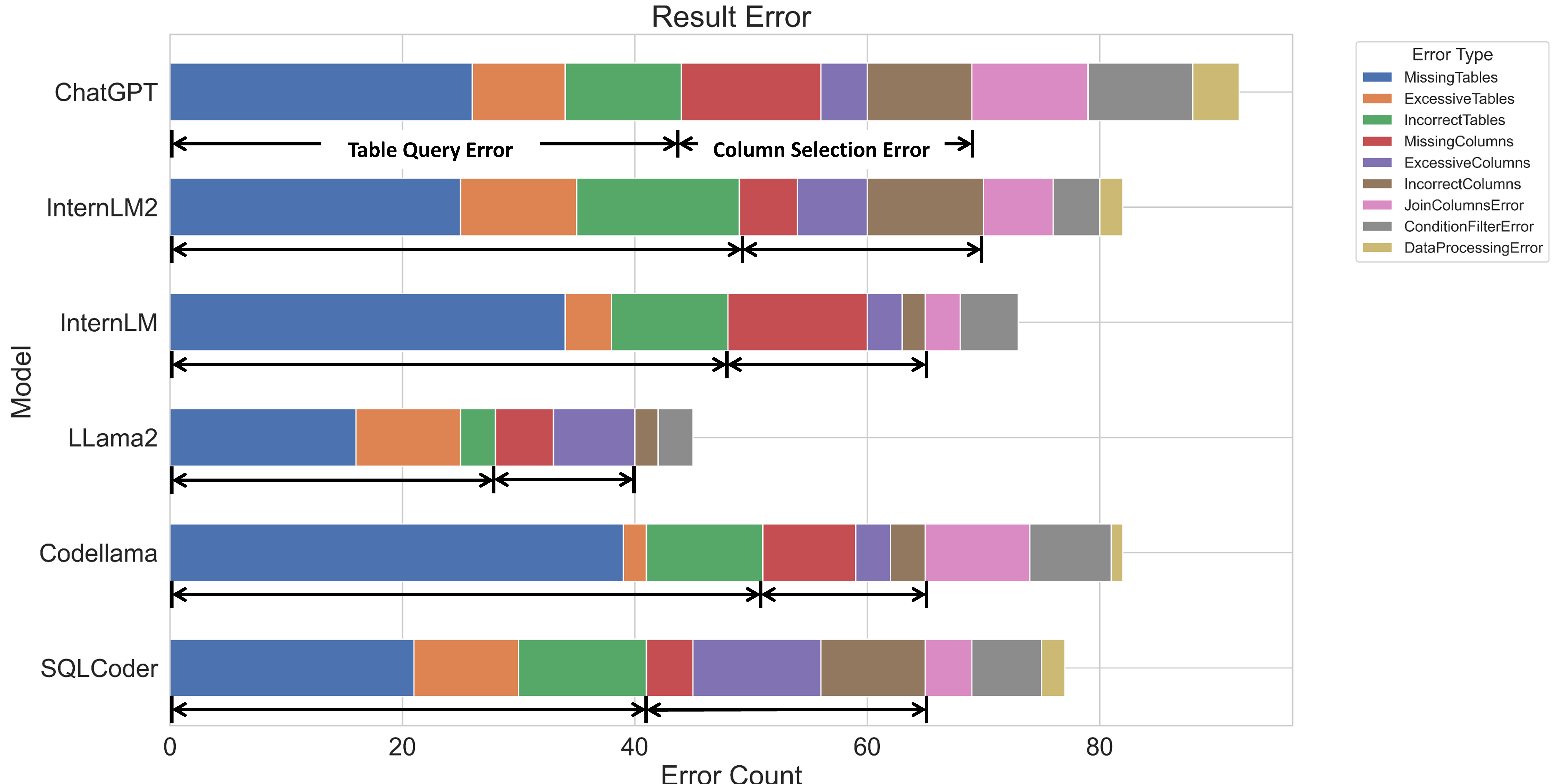}
    \vspace{-8pt}
    \caption{Result Errors for incorrect SQL statements generated by LLMs are categorized into five types: Table Query Errors, Column Selection Errors, Join Column Errors, Condition Filter Errors, and Data Processing Errors. Specifically, Table Query Errors encompass issues related to excessive, missing, or incorrect tables, while Column Selection Errors include excessive, missing, or incorrect columns.}
    \label{fig:result_error}
    \vspace{-10pt}
\end{figure}

Detailed comments for all Result Error are shown in Table~\ref{tab:debug_comments}.
In addition, Figure~\ref{fig:all_error} and Figure~\ref{fig:result_error} provide more detailed information about error types, including the quantity of erroneous data.

\begin{table*}[htbp]
\centering
  \caption{Comments for different categories of Result Error used in SQL debugging.}
  \label{tab:debug_comments}
  \resizebox{0.95\textwidth}{!}{%
\begin{tabular}{@{}lll@{}}
\toprule
\textbf{Error Type} &
  \textbf{Subcategory} &
  \multicolumn{1}{c}{\textbf{comments prompt}} \\ \midrule
\multirow{4}{*}{\textbf{Table Query Error}} &
  \textbf{\begin{tabular}[c]{@{}l@{}}Excessive \\ Tables\end{tabular}} &
  The tables you inquired about is incorrect, you query too much tables. \\ \cmidrule(l){2-3} 
 &
  \textbf{\begin{tabular}[c]{@{}l@{}}Missing \\ Tables\end{tabular}} &
  The tables you inquired about is incorrect, you need to query more tables. \\ \cmidrule(l){2-3} 
 &
  \multirow{2}{*}{\textbf{\begin{tabular}[c]{@{}l@{}}Incorrect \\ Tables\end{tabular}}} &
  \multirow{2}{*}{The tables you inquired about is incorrect.} \\
 &
   &
   \\ \midrule
\multirow{4}{*}{\textbf{Column Selection Error}} &
  \textbf{\begin{tabular}[c]{@{}l@{}}Excessive \\ Columns\end{tabular}} &
  \begin{tabular}[c]{@{}l@{}}You have found the correct tables.\\ But you select wrong columns,you select too much Columns.\end{tabular} \\ \cmidrule(l){2-3} 
 &
  \textbf{\begin{tabular}[c]{@{}l@{}}Missing \\ Columns\end{tabular}} &
  \begin{tabular}[c]{@{}l@{}}You have found the correct tables.\\ But you select wrong columns,you need to select more Columns.\end{tabular} \\ \cmidrule(l){2-3} 
 &
  \multirow{2}{*}{\textbf{\begin{tabular}[c]{@{}l@{}}Incorrect \\ Columns\end{tabular}}} &
  \multirow{2}{*}{You have found the correct tables.But you select wrong columns.} \\
 &
   &
   \\ \midrule
\multicolumn{2}{l}{\textbf{Join Columns Error}} &
  \begin{tabular}[c]{@{}l@{}}You have found the correct tables.\\ You have selected the correct Columns.\\ But you combine wrong rows when JOIN two tables.\end{tabular} \\ \midrule
\multicolumn{2}{l}{\textbf{Condition Filter Error}} &
  \begin{tabular}[c]{@{}l@{}}You have found the correct tables.You have selected the correct Columns.\\ You have combined (JOIN) the correct tables.\\ But an error occurred in the conditional filter.\end{tabular} \\ \midrule
\multicolumn{2}{l}{\textbf{Data Processing Error}} &
  \begin{tabular}[c]{@{}l@{}}You have found the correct tables.\\ You have selected the correct Columns.\\ You have combined (JOIN) the correct tables.\\ You have used the correct conditional filtering. \\ But there was an error in your processing of the data.\end{tabular} \\ \bottomrule
\end{tabular}%
}
\end{table*}

\subsection{Details for SQL Optimization}
We suspect the reasons LLMs are challenging to achieve effective SQL optimization are as follows:
\begin{itemize}
    \item \textbf{Lack of in-context learning guidance.} For self-debug, models can benefit from the error messages given by DBMS (see the last two rows of Figure~\ref{fig:self-debug}). However, SQL optimization tasks lack explicit feedback from the system and guidance through In-context learning, despite the comprehensively designed prompt templates.
    \item \textbf{Lack of pre-training data.} During the pre-training phase, models can access massive correct SQL statements as training data, yet they may lack information on what kind of SQL statements are efficient.
    \item \textbf{System-level optimization is inherently challenging.} The efficiency of a SQL query can only be assessed by executing it within a database system. This is a system-level optimization beyond the scope of syntactic-level and semantic-level language modeling areas in which LLMs excel. It is challenging for models to capture the mapping between SQLs and their execution efficiency.
\end{itemize}

\subsection{Details for SQL-to-Text}
 
Detailed description of evaluation metrics used for SQL-to-Text task are as follows:
\begin{itemize}
    \item \textbf{\textit{Rouge}}~\cite{rouge} focuses on token-level syntactical similarity by breaking down the predicted question and the GT question into N-grams and calculating the recall rate. We use ROUGE-1, ROUGE-2, and ROUGE-L, respectively.
    \item \textbf{\textit{BertScore}}~\cite{bertscore} focuses on semantic similarity between predicted and GT questions. Based on the BERT pre-trained model, BertScore uses contextual embeddings to describe sentences and calculates the cosine similarity between them.
    \item \textbf{\textit{Using LLMs as evaluator}} is an intuitive and recently prevalent evaluation approach. This involves presenting two questions to an LLM to determine whether they are semantically identical, with the accuracy rate being recorded. We use InternLM2 as evaluator, which achieves the best performance in the first two metrics. 
    In practice, to mitigate the potential bias introduced by positional preference, we alter the order of predicted and GT questions in the input of InternLM2 and report the average result of two experiments. 
    
\end{itemize}

\section{Prompt Template}

\subsection{Text-to-SQL}
\label{app:text-to-sql prompt_template}

Listing~\ref{lst:DDL-HTML}-~\ref{lst:SimpleDDL-Coding-Chat} demonstrate eight prompt templates evaluated in the end-to-end Text-TO-SQL task, where ``SimpleDDL-MD-Chat'' obtains the best evaluation performance.

\begin{lstlisting}[language=Special_Prompt, label={lst:DDL-HTML}, caption={Prompt template ``DDL-HTML-Chat/Complete''. The content in parentheses is unique to``Complete''.}, float=ht]
Figure out corresponding SQLite SQL Query of Question according to database.
<Database>
&&CREATE TABLE stadium (stadium_id NUMBER PRIMARY KEY, ...); 
CREATE TABLE singer (singer_id NUMBER PRIMARY KEY, name TEXT, ...);&& 
</Database>
<Question>^^How many singers do we have?^^</Question>
(<SQL> @@SELECT@@)
\end{lstlisting}
\begin{lstlisting}[language=Special_Prompt, label={lst:SimpleDDL-HTML}, caption={Prompt template ``SimpleDDL-HTML-Chat/Complete''. The content in parentheses is unique to``Complete''.}, float=ht]
Figure out corresponding SQLite SQL Query of Question according to database.
<Database>
&&stadium(Stadium_ID,Location,Name,Capacity,Highest,Lowest,Average);
singer(Singer_ID,Name,Country,Song_Name,...);&&
</Database>
<Question>^^How many singers do we have?^^</Question>
(<SQL> @@SELECT@@)
\end{lstlisting}

\begin{lstlisting}[language=Special_Prompt, label={lst:DDL-MD}, caption={Prompt template ``DDL-MD-Chat/Complete''. The content in parentheses is unique to``Complete''.}, float=ht]
### Answer the question by sqlite SQL query only and with no explanation
### Sqlite SQL tables, with their properties:
&&CREATE TABLE stadium (stadium_id NUMBER PRIMARY KEY,...); 
CREATE TABLE singer (singer_id NUMBER PRIMARY KEY, name TEXT, ...);&& 
### Question: ^^How many singers do we have?^^
### SQL: (@@SELECT@@) 
\end{lstlisting}

\begin{lstlisting}[language=Special_Prompt, label={lst:SimpleDDL-MD}, caption={Prompt template ``SimpleDDL-MD-Chat/Complete''. The content in parentheses is unique to``Complete''.}, float=htbp]
### Answer the question by sqlite SQL query only and with no explanation
### Sqlite SQL tables, with their properties:
#
# &&stadium(Stadium_ID,Location,Name,Capacity,Highest,Lowest,Average);&&
# &&singer(Singer_ID,Name,Country,Song_Name,...);&&
#
### ^^How many singers do we have?^^
### SQL: (@@SELECT@@) 
\end{lstlisting}

\begin{lstlisting}[language=Special_Prompt, label={lst:DDL-Coding}, caption={Prompt template ``DDL-Coding-Chat/Complete''. The content in parentheses is unique to``Complete''.}, float=htbp]
/* Given the following database schema: */
&&CREATE TABLE stadium (stadium_id NUMBER PRIMARY KEY, ...);

CREATE TABLE singer (singer_id NUMBER PRIMARY KEY, name TEXT, ...);&&

/* Answer the following by SQLite SQL Query according to database: ^^Show name, country, age for all singers ordered by age from the oldest to the youngest.^^ */
/* SQL Query here*/
(@@SELECT@@) 
\end{lstlisting}

\begin{lstlisting}[language=Special_Prompt, label={lst:SimpleDDL-Coding-Chat}, caption={Prompt template ``SimpleDDL-Coding-Chat/Complete''. The content in parentheses is unique to``Complete''.}, float=htbp]
/* Given the following database schema: */
&&stadium(Stadium_ID,Location,Name,Capacity,Highest,Lowest,Average);
singer(Singer_ID,Name,Country,Song_Name,...);&&

/* Answer the following by SQLite SQL Query according to database: ^^Show name, country, age for all singers ordered by age from the oldest to the youngest.^^ */
/* SQL Query here*/
(@@SELECT@@) 
\end{lstlisting}

\subsection{SQL Debugging}
\label{app:Debugging prompt_template}

Listing~\ref{lst:debug_error_classification} is used for error classification. In the detailed categorization process of Result Error, Table Query Error, Column Selection Error, and Join Column Error can be determined by comparing SQL Query with the ground truth using predefined rules. For Data processing Error and Condition Filter Error, we employ InternLM2-20B for binary classification tasks.
Listing~\ref{lst:debug_prompt} is the prompt template used for SQL debugging tasks.

\begin{lstlisting}[language=Special_Prompt, label={lst:debug_error_classification}, caption={Prompt template for conducting the ``Condition Filter Error'' and ``Data Processing Error'' classification tasks.}, float=htbp]
You are an expert in SQL queries. Please provide the error categories for incorrect SQL queries based on the Question and the correct SQL query.
Please think step by step and check for the following errors in order:
1. Condition Filter Error: Incorrect filtering of conditions. 
2. Data Processing Error:The condition is filtered correctly, but the data processing is wrong. Note that the premise of this error is that the conditional filtering is correct.

Question: ^^What is the name and capacity for the stadium with highest average attendance?^^
Correct SQL Query: @@SELECT name ,  capacity FROM stadium ORDER BY average DESC LIMIT 1@@
Wrong SQL Query: @@SELECT Name, Capacity FROM stadium WHERE Average = (SELECT MAX(Average) FROM stadium) ORDER BY Highest DESC@@

Give your Thought and Answer based on the information above.
\end{lstlisting}

\begin{lstlisting}[language=Special_Prompt, label={lst:debug_prompt}, caption={Prompt template ``w/ Wrong SQL + All\_error\_info + Comment'' for SQL Debugging.}, float=htbp]
### Write the correct SQLite SQL Query corresponding to the Question based on the database, the Wrong SQL Query and the cause of the error.
### Sqlite SQL tables, with their properties:
#
#{SimpleDDL}
#
### Question: ^^Under whose administration does the school with the highest number of test takers whose total SAT Scores are greater or equal to 1500 belong to? Indicate his or her full name.^^
### Wrong SQL Query:
@@SELECT T1.AdmFName1 ,  T1.AdmLName1 FROM schools AS T1 JOIN satscores AS T2 ON T1.CDSCode = T2.cds WHERE T2.NumTstTakr = ( SELECT NumTstTakr FROM satscores GROUP BY cds HAVING NumGE1500  >=  1500 ORDER BY NumTstTakr DESC LIMIT 1 ) @@
### Error Information:
Executed correctly, but with the wrong result.
You have found the correct tables. But you select wrong columns, you need to select more Columns.
### Correct SQL: 
\end{lstlisting}

\subsection{SQL Optimization}
\label{app:Optimization prompt_template}

Listing~\ref{lst:opt_w/ demo + comments} is the prompt template ``w/ demo + comments'' used for SQL optimization tasks. Other variations of the template correspond to the removal of relevant content.

\begin{lstlisting}[language=Special_Prompt, label={lst:opt_w/ demo + comments}, caption={Prompt template ``w/ demo + comments'' for SQL Optimization.}, float=htbp]
### Rewrite and optimize the given SQL query to improve SQL query efficiency and minimize SQL execution time while ensuring correctness. Only output sql query, do not output any other content.Only output sql query, do not output any other content.
### Here are some reference cases:
#
# Question: ^^List out the age of users who located in Vienna, Austria obtained the badge?^^
# SQL Query: @@SELECT Age FROM users WHERE Location = 'Vienna, Austria' AND Id IN (SELECT UserId FROM badges)@@
# New SQL Query: @@SELECT u.Age FROM users AS u INNER JOIN badges AS b ON u.Id = b.UserId WHERE u.Location = 'Vienna, Austria'@@
# Explanation: By applying a JOIN operation instead of a subquery with IN can improve efficiency, as the database may execute the JOIN and filtering processes concurrently in just one operation without the need to store the intermediate results to filter primary query.
#...
#
### Sqlite SQL tables, with their properties:
#
#{SimpleDDLdd}
#
### Question: ^^What is the short name and fifa ID for Queens Park Rangers Football Team?...^^
### SQL Query:@@SELECT team_short_name ,  team_fifa_api_id FROM Team WHERE team_long_name = "Queens Park Rangers"@@
### New SQL Query: 
\end{lstlisting}

\subsection{SQL-to-Text}
\label{app:sql2text}

Listing~\ref{lst:sql_to_text} is the prompt template used for SQL-to-Text task. Listing~\ref{lst:Semantic_consistency} is utilized to assess the consistency between the transformed text description and the original user query semantics. To mitigate the influence of the order of the two text inputs, we employ ChatGPT and InternLM2-20B to conduct two sequential tests and calculate the average of the results.

\begin{lstlisting}[language=Special_Prompt, label={lst:sql_to_text}, caption={Prompt template  for SQL-to-Text.}, float=htbp]
<Instruction>
You are an expert in database analysis and processing of SQL statements.
I will provide an SQL statement and relevant evidence. You need to help me analyze what problem this SQL statement is solving.
Here are some reference cases:
SQL:@@SELECT list_id FROM lists_users WHERE user_id = 85981819 ORDER BY list_creation_date_utc ASC LIMIT 1@@
question:What is the list ID that was first created by user 85981819?
SQL:@@SELECT COUNT(T2.user_id) FROM movies AS T1 INNER JOIN ratings AS T2 ON T1.movie_id = T2.movie_id WHERE T1.movie_title = 'Pavee Lackeen: The Traveller Girl' AND T2.rating_score = 4@@
question:^^How many users gave \"Pavee Lackeen: The Traveller Girl\" movie a rating score of 4?^^
Please answer strictly in the following format and do not change the format arbitrarily:
question:This is a problem description.
</Instruction>
<SQL>@@SELECT ...@@</SQL>
\end{lstlisting}

\begin{lstlisting}[language=Special_Prompt, label={lst:Semantic_consistency}, caption={Prompt template  for semantic consistency checking.}, float=htbp]
<Instruction>Determine whether the following two sentences ask the same question and whether their corresponding answers are the same.</Instruction>
<sentence1>^^How many singers do we have?^^</sentence1>
<sentence2>^^How many singers are there in total?^^</sentence2>
<Question>Just output True or False, do not output anything else</Question>
\end{lstlisting}

\subsection{Schema Linking}
\label{app:schema_linking}
Listing~\ref{lst:zero_shot_linking} is a schema linking prompt template proposed in C3~\cite{c3}. Listing~\ref{lst:few_shot_linking} is used for few-shot learning proposed in DIN-SQL~\cite{din-sql}. In addition, we conduct tests on the PreSQL method. This method generates an SQL query statement based on the user query and database schema, and extracts the corresponding tables from that statement.

\begin{lstlisting}[language=Special_Prompt, label={lst:zero_shot_linking}, caption={Prompt template for zero-shot schema linking.}, float=htbp]
Given the database schema and question, perform the following actions: 
1 - Rank all the tables based on the possibility of being used in the SQL according to the question from the most relevant to the least relevant, Table or its column that matches more with the question words is highly relevant and must be placed ahead.
2 - Check whether you consider all the tables.
3 - Output a list object in the order of step 2, Your output should contain all the tables. The format should be like: 
[
    "table_1", "table_2", ...
]

Database schemas with their properties:
{SimpleDDL}

Question: ^^What is the short name and fifa ID for Queens Park Rangers Football Team?In the database, short name of the football team refers to team_short_name; Queens Park Rangers refers to team_long_name = 'Queens Park Rangers';fifa ID refers to team_fifa_api_id.^^
Answer(Only output the list object containing all tables, do not output other content): 
\end{lstlisting}

\clearpage
\begin{lstlisting}[language=Special_Prompt, label={lst:few_shot_linking}, caption={Prompt template for few-shot schema linking.}, float=htbp]
Given the database schema and question, perform the following actions: 
1 - Evaluate the importance of each table **in relation to the SQL query**, prioritizing tables and columns that closely match the question words. Rank the tables from the most crucial to the least crucial.
2 - Focus on identifying and listing only the most important tables based on the evaluation in step 1.
3 - Output a list object representing the order determined in step 2. The output should include **the most important tables** and follow this format: 
[
    "most_important_table_1", "most_important_table_2", ...
]

Schema:
# &&department(Department_ID,Name,Creation, ...)&&
# &&head(head_ID,name,born_state,age)&&
# &&management(department_ID,head_ID,temporary_acting)&&
Foreign key:
&&management(department_ID) REFERENCES department(Department_ID)\nmanagement(head_ID) REFERENCES head(head_ID)&&
Question: ^^what are the distinct creation years of the departments managed by a secretary born in state 'Alabama'?^^
Answer: ["department","management","head"]

Schema:
# &&Country(id,name)&&
# &&League(id,country_id,name)&&
# &&Player(id,player_api_id,player_name,player_fifa_api_id,birthday,height,weight)&&
# &&Player_Attributes(id,player_fifa_api_id, ...)&&
# &&Team(id,team_api_id,team_fifa_api_id,team_long_name,team_short_name)&&
# &&Team_Attributes(id,team_fifa_api_id, ...)&&
# &&sqlite_sequence(name,seq)&&
Foreign key:
&&Player_Attributes(player_api_id) REFERENCES Player(player_api_id)\nLeague(country_id) REFERENCES country(id)\nTeam_Attributes(team_api_id) REFERENCES Team(team_api_id)\nMatch(away_player_11) REFERENCES Player(player_api_id)&&
Question: ^^List the names of all left-footed players who have overall rating between 85 and 90.^^
Answer: ["Player","Player_Attributes"]

Database schemas with their properties:
{SimpleDDL}

Question: ^^What is the short name and fifa ID for Queens Park Rangers Football Team? In the database, short name of the football team refers to team_short_name; Queens Park Rangers refers to team_long_name = 'Queens Park Rangers';fifa ID refers to team_fifa_api_id.^^
Answer: 
\end{lstlisting}

\end{document}